\documentclass[sigconf]{acmart}
\usepackage{multirow}
\usepackage{makecell}
\usepackage{enumitem}
\usepackage{titlepic}

\usepackage{algorithm}
\usepackage{algpseudocode}
\usepackage{amsmath}
\usepackage{amsfonts}
\usepackage{booktabs}
\usepackage{subcaption}
\usepackage{bbding}
\usepackage{multicol}
\usepackage{graphicx}
\usepackage{color}
\AtBeginDocument{%
  \providecommand\BibTeX{{%
    \normalfont B\kern-0.5em{\scshape i\kern-0.25em b}\kern-0.8em\TeX}}}

\renewcommand\footnotetextcopyrightpermission[1]{}
\begin{document}

 \title{\includegraphics[height=25px]{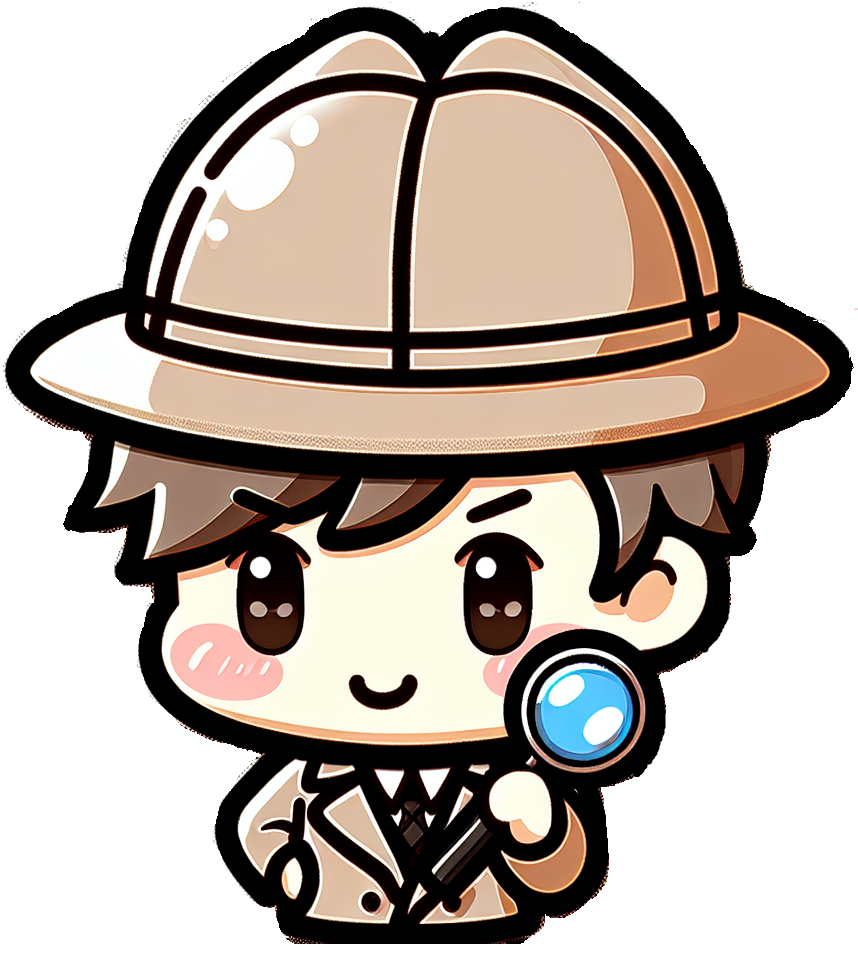}Omni-IML: Towards Unified Image Manipulation Localization}

% \affiliation{%

\author{Chenfan Qu}
\affiliation{%
  \institution{South China University of Technology}  
  \city{Guangzhou}
  \country{China}
}

\author{Yiwu Zhong}
\affiliation{%
  \institution{The Chinese University of Hong Kong}
  \city{Hong Kong}
  \country{China}
}

\author{Fengjun Guo}
\affiliation{%
  \institution{Intsig Information Co., Ltd}
  \city{Shanghai}
  \country{China}
}

\author{Lianwen Jin}
\affiliation{%
  \institution{South China University of Technology}
  \city{Guangzhou}
  \country{China}
}

\begin{abstract}
  Existing Image Manipulation Localization (IML) methods mostly rely heavily on task-specific designs, making them perform well only on the target IML task, while joint training on multiple IML tasks causes significant performance degradation, hindering real applications.
  To this end, we propose \textbf{Omni-IML}, the first generalist model designed to unify IML across diverse tasks. 
  Specifically, Omni-IML achieves generalization through three key components: (1) a \textbf{Modal Gate Encoder}, which adaptively selects the optimal encoding modality per sample, (2) a \textbf{Dynamic Weight Decoder}, which dynamically adjusts decoder filters to the task at hand, and (3) an \textbf{Anomaly Enhancement} module that leverages box supervision to highlight the tampered regions and facilitate the learning of task-agnostic features. 
  Beyond localization, to support interpretation of the tampered images, we construct \textbf{Omni-273k}, a large high-quality dataset that includes natural language descriptions of tampered artifact. It is annotated through our automatic, chain-of-thoughts annotation technique.
  We also design a simple-yet-effective interpretation module to better utilize these descriptive annotations.
  Our extensive experiments show that our single Omni-IML model achieves state-of-the-art performance across all four major IML tasks, providing a valuable solution for practical deployment and a promising direction of generalist models in image forensics. Our code and dataset will be publicly available.
\end{abstract}



\maketitle

\section{Introduction}
\label{sec:intro}

The rapid advancement of image processing technique makes it easy to manipulate images, incurring risks in social media security~\cite{sun2023safl}. Despite the progress made in recent years, existing IML models are designed for individual IML tasks on specific image types (e.g., natural image or document etc.) and usually fall short on other IML tasks. 
Consequently, the maintenance costs are high since every IML task requires an independent, task-specific IML model.

A naive solution is to jointly train an IML model using the data of all available IML tasks. 
However, joint training usually leads to an obvious performance degradation on all IML tasks, making the predictions unreliable. For example, HiFi-Net~\cite{hifi_net_xiaoguo} suffers from joint training and thus uses two different sets of model parameters for natural image IML and face IML, respectively. 
There are two main reasons why existing IML methods still cannot provide generalization across IML tasks after joint training:

First, existing IML methods heavily rely on task-dependent architecture designs, input modalities, and training strategies to detect tampering clues. These designs work well for the target IML task, but usually fall short on other IML tasks. For example, edge anomaly enhancement modules~\cite{dong2022mvss, Yu_2024_CVPR} and object attention modules~\cite{wang2022objectformer, Li_2024_CVPR} have made significant progress in identifying forged natural objects. However, they can hardly work well on document images where edge artifacts are less obvious and object features are not distinct. Early frequency-vision~\cite{CVPR2023DocTamper} fusion performs well on document images but has obvious performance degradation on natural images that cover much more noise and diversity.

Second, existing IML methods lack the design to distinguish diverse tampering features across different IML tasks. The IML task is challenging since the tampering methods are diverse and produce different subtle tampering cues. Further, it is even more challenging to handle various IML tasks using a unified model.

\begin{figure}[t!]
  \centering
  \includegraphics[width=\linewidth]{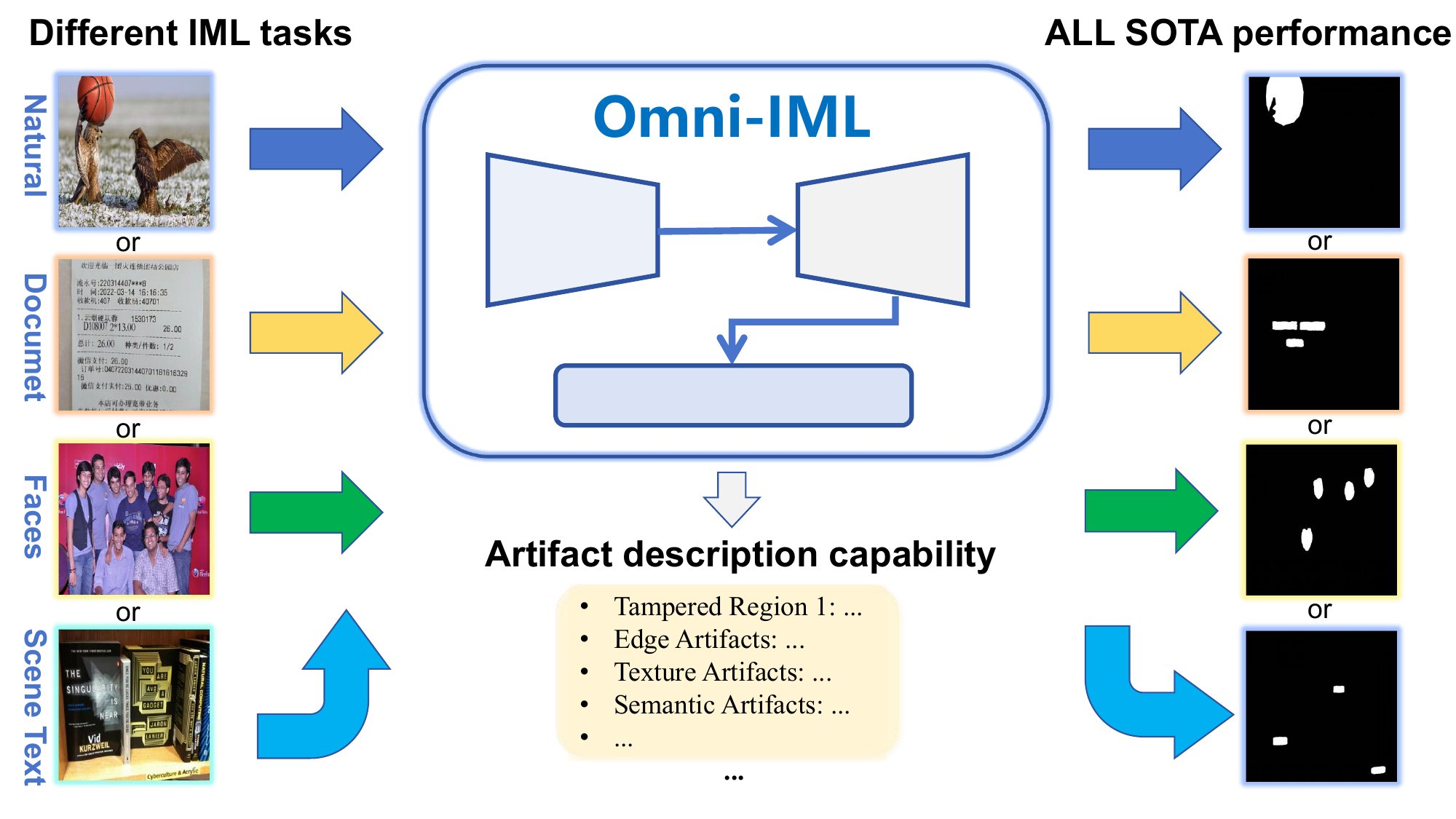}
  \vspace{-0.8cm}
  \caption{The proposed Omni-IML is the first generalist model for IML. A single model can simultaneously achieve state-of-the-art performance on multiple major IML tasks, without task-specific and benchmark-specific fine-tuning.}
  \vspace{-0.4cm}
  \label{fig: teaser}
\end{figure}

To address the above issues, we propose Omni-IML, the first generalist model that can simultaneously perform well on multiple major IML tasks, as shown in Fig~\ref{fig: teaser}. 
Specifically, a Modal Gate Encoder is proposed to automatically select the optimal encoding modality for each input sample, based on the characteristics of the input image. And a Dynamic Weight Decoder is proposed to adaptively select the optimal decoder filters for each sample, assisting the generalist model to better cope with the highly diverse tampering features from multiple image types. 
These sample-adaptive designs can provide flexibility so that the model can adapt to each input sample and achieve generalization. Further, an Anomaly Enhancement module is introduced to enhance the features of tampered regions with a novel box supervision design. 

Merely localizing tampered regions is not convincing enough~\cite{xu2024fakeshield, huang2024ffaa}. We also seek to improve model reliability by enabling the language description on both visual and semantic artifacts of the tampered image. 
Despite the recent progress in interpretable IML on natural images, there still lacks public data for interpretable IML on documents, scene text images and uncut deepfake images. 
The interpretable IML annotation generation pipelines with GPT-4o in previous works~\cite{xu2024fakeshield, huang2024sida} work well on natural images, but do not work well on documents, uncut deepfakes and scene texts, where a tampered image usually contains multiple tampered regions and the image artifact are less obvious~\cite{CVPR2023DocTamper}. 
There are two main reasons behind this gap: First, previous methods obtain the content, position and artifact descriptions of the tampered object with only a single query to GPT-4o. When there are multiple tampered objects in one image, the GPT-4o is easily distracted and messes up the descriptions among different targets. Second, when the image artifact is less obvious, the GPT-4o's response will be less confident and even incorrect, resulting in low-quality annotations.

To this end, we propose a novel chain-of-thoughts pipeline, which solves the above annotation issues through step-by-step focused analysis and self-examination. Specifically, we first recognize all the tampered objects and their locations. For each tampered object, we query GPT-4o to generate its artifact description with an elaborate prompt, and then query GPT-4o to examine its previous response. With the proposed method, we generate artifact descriptions on forged natural images, documents, faces, scene texts, and construct a large-scale, comprehensive, high quality dataset Omni-273k. To make better use of our Omni-273k, we further propose a simple-yet-effective interpretation module that improves model's artifact description performance through a reference visual prompt.

Due to limited computational resources, it is not practical to validate our model on all IML tasks and datasets in the world. Instead, we validated our methods on hundreds of representative IML tasks (e.g., IML on certificates, product photos, artwork, street photos, cards, signs, group photos, receipts, etc.), which can be categorized into four distinct \textbf{major} IML tasks, including natural image IML, document IML, face IML and scene text IML, \textbf{covering the vast majority of the recent IML research}. 
Extensive experiments on the four major IML tasks show that joint training the existing IML methods on all tasks leads to significant performance degradation and the trained IML models are inadequate to handle multiple IML tasks simultaneously. While our Omni-IML can minimize the performance degradation and simultaneously achieve state-of-the-art performance across all the four major IML tasks, demonstrating high scalability and effectiveness.

In summary, our main contributions are as follows:
\begin{itemize}
    \item We propose Omni-IML, \textbf{the first IML generalist model} that enables generalizable image forgery localization and interpretation, serving as a pioneering effort in the field.
    \item Unified IML modeling is achieved by multliple \textbf{novel and effective modules}, consisting of Modal Gate Encoder, Anomaly Enhancement, and Dynamic Weight Decoder.
    \item Further, to interpret image artifact in natural language, we propose a novel chain-of-thoughts annotation technique to automatically construct Omni-273k, a large-scale and high-quality dataset. A simple-yet-effective interpretation module is also proposed to better leverage our data. 
    \item Extensive experiments demonstrate that our \textbf{single generalist model} can simultaneously achieve \textbf{state-of-the-art} results across different major IML tasks.
\end{itemize}

\vspace{-0.2cm}
\section{Related works}
%-------------------------------------------------------------------------
% \subsection{Specialized Image Manipulation Localization}
\noindent \textbf{Natural IML} aims to identify the tampered regions in daily-life style images. MVSS-Net~\cite{dong2022mvss} introduces ESB module to enhance edge inconsistency. ObjectFormer~\cite{wang2022objectformer} proposes an object encoder to learn object-level attention for better feature extraction. Trufor~\cite{guillaro2023trufor} benefits from the Bayar noise filters. These models perform well on natural images, but mostly not well enough on other image types (e.g. documents), due to the the absence of natural object, edge artifact and noise artifact in these scenarios. FakeShield~\cite{xu2024fakeshield} improves natural IML based on LISA~\cite{lai2024lisa}, which is constrained in multi-target scenarios~\cite{xia2024gsva} and is thus challenged to achieve reliable IML on text images and uncut deepfake images~\cite{le2021openforensics}.

\noindent \textbf{Document IML} aims to localize the forged regions in document images. Early works~\cite{doc1, doc2} achieve document forensics through template-matching based methods. These methods work well on clean documents but struggle with complex images such as photographed documents. DTD~\cite{CVPR2023DocTamper} improves document IML through early fusion of vision and frequency features. However, the model will be seriously distorted in many cases of natural and face images where the frequency features are too noisy. TIFDM~\cite{dong2024robust} proposes high-level spatial attention to suppress the false alarms in documents, but it is limited on complex natural images.

\noindent \textbf{Face IML} aims to localize AI-generated fake faces. HiFiNet~\cite{hifi_net_xiaoguo} utilizes metric learning for better texture anomaly capturing. MoNFAP~\cite{liu2024hfnet} improves texture anomaly detection with a set of noise filters. These methods show generalization on face IML but are suboptimal on natural and document images, where the tampered regions are small in size and the texture anomalies are less obvious.

\noindent \textbf{Scene Text IML} aims to localize tampered natural scene text in arbitrary styles and complex backgrounds. Previous works relies on specific noise patterns~\cite{wang2022tic} or pre-training on authentic scene text images, limiting their generalization across other IML tasks.

\begin{figure*}[t!]
  \centering
  \includegraphics[width=\linewidth]{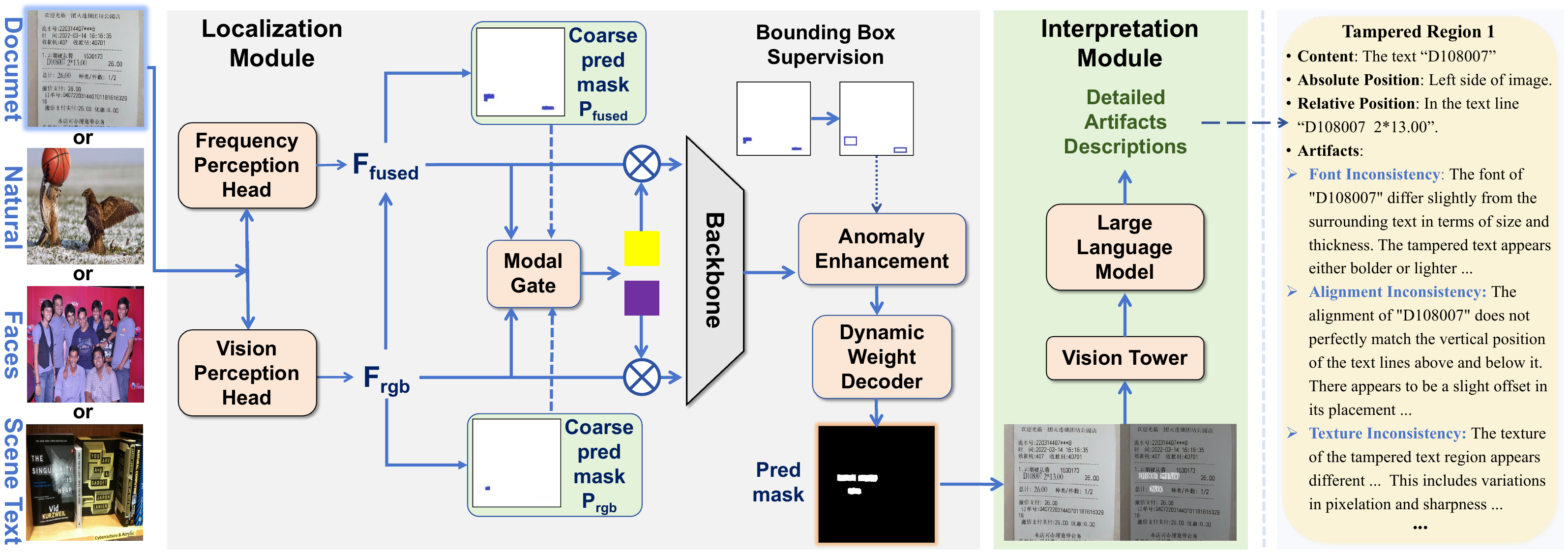}
  \vspace{-0.4cm}
  \caption{The overall framework of the proposed Omni-IML.}
  \label{fig: main}
  \vspace{-0.2cm}
\end{figure*}

\begin{figure*}[t!]
  \centering
  \includegraphics[width=\linewidth]{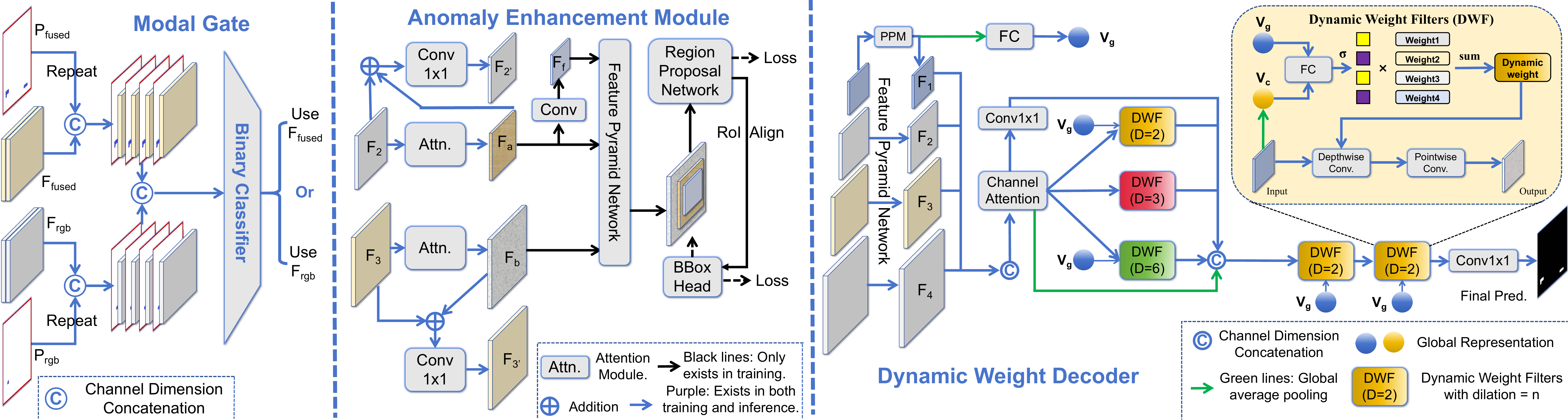}
  \vspace{-0.4cm}
  \caption{The proposed Modal Gate (left), Anomaly Enhancement Module (middle) and Dynamic Weight Decoder (right).}
  \label{fig: Modules}
  \vspace{-0.3cm}
\end{figure*}

\section{Methodology}
As shown in Figure~\ref{fig: main}, the Omni-IML has a localization module to identify whether the input image is tampered with and to localize the tampered region, and an interpretation module to describe the artifact for tampered image in natural language.

The \textbf{localization module} is roughly based on an encoder-decoder architecture. The \textbf{Modal Gate Encoder} consists of: (1) Visual Perception Head to extract visual features from the original images; (2) Frequency Perception Head to extract frequency features; (3) a Modal Gate to automatically determine the optimal modality for the following encoding process; (4) a backbone model to extract multi-scale high-level features. The \textbf{Dynamic Weight Decoder} adaptively selects the sample-wise optimal decoder filters and outputs the final mask prediction. We also design an \textbf{Anomaly Enhancement} module between the encoder and decoder, to enhance the common features of tampered regions from various image types. 

The \textbf{interpretation module} is a multimodal LLM, it describes the image artifact by examining both the original input image and a reference visual prompt constructed by highlighting the decoder's mask prediction on the original input image.

\subsection{Modal Gate Encoder}
\noindent \textbf{Key Idea.} The frequency feature is a double-edged sword for the IML generalist: it can help to detect visually consistent tampering in some cases, but it can also degrade the model performance when the image is too low-quality and the frequency is too noisy. 
As a result, neither pure vision nor vision+frequency modeling can consistently provide the optimal solution. To achieve general IML through a flexible encoding modality, we propose the Modal Gate, which automatically determines the optimal encoding modality (frequency+vision or pure vision) for each input sample. The key idea of our Modal Gate Encoder is to \textbf{automatically identify the optimal modality by analyzing whether the frequency features contain too much noise, and which coarse prediction seems more confident, reliable, and accurate.}

\noindent \textbf{Image Encoding}. As shown in Figure~\ref{fig: main}, we extract vision features $F_{rgb}$ using Vision Perception Head, extract frequency features using Frequency Perception Head, and obtain the fused features $F_{fused}$ by fusing the vision and frequency features with a conv-layer. Both the two perception heads consist of several conv-layers, with the same structure as those in the DTD model~\cite{CVPR2023DocTamper}. Two coarse mask predictions $P_{rgb}$ and $P_{fused}$ are obtained from $F_{rgb}$ and $F_{fused}$ with two conv-layers respectively.

\noindent \textbf{Modal Gate.} As shown in the left part of Figure~\ref{fig: Modules}, $F_{rgb}$, $F_{fused}$, $P_{rgb}$ and $P_{fused}$ are channel-dimension concatenated and fed into the Modal Gate for optimal modality prediction. The Modal Gate is a binary classifier consists of several conv-layers and activation layers to determine whether to use $F_{fused}$ or $F_{rgb}$ as the encoder input, {by observing the noise level and the confidence} of $F_{fused}$, $F_{rgb}$ and their corresponding coarse predictions $P_{rgb}$ and $P_{fused}$. More details are presented in the appendix.

The Modal Gate Encoder maximizes the advantages of frequency domain modeling especially when the visual anomalies are limited (e.g. document images), and avoids its drawbacks when the image is too complex and noisy (e.g. natural images). Our Modal Gate Encoder extracts the best features from different image types and thus considerably benefits the generalist IML model.

\subsection{Anomaly Enhancement}
\noindent \textbf{Key Idea.} Different image types from different IML tasks produce different features, joint training brings much more noise to the features and confuses IML model. To tackle this, we propose to enhance the contrast of forged regions and improve the feature extraction across diverse image types through including an extra box supervision during training. 

However, directly training the model with the both detection and segmentation frameworks may also cause task competition for model parameters~\cite{heuer2021multitask} and weaken model performance, while directly scaling up the model parameters could alleviate the competition but will increase computation burden. To address this, we propose a novel collaboration module Anomaly Enhancement (AE).

\noindent \textbf{Method}. As shown in the middle of Figure~\ref{fig: Modules}, for the input features $F_{2}$ and $F_{3}$ in $\frac{1}{8}$ and $\frac{1}{16}$ input image size, we first extract task-agnostic features $F_{a}$ and $F_{b}$ with two attention layers that used to decouple and to minimize negative impact from the segmentation supervision.
After that, $F_{a}$ and $F_{b}$ are processed by the detection modules, including two Feature Pyramid Networks (FPNs) ~\cite{lin2017fpn} and the Faster R-CNN's~\cite{ren2015fasterrcnn} Region Proposal Network (RPN) and box head. The detection modules (black arrows in the middle of Figure~\ref{fig: Modules}) are only present during training. Including the two cascaded FPNs reduces parameter competition from the detection framework and discarding them during inference ensures the computation efficiency, successfully addressing the dilemma. The AE module is trained in an end-to-end manner with the same four loss functions as the Faster R-CNN, including the two classification losses and two regression losses for RPN and box head respectively. After training, the $F_{a}$ and $F_{b}$ contain positive features enhanced by the detection supervision, we add them to the original features $F_{2}$ and $F_{3}$ and fuse them with conv-layer to get $F_{2'}$ and $F_{3'}$. More details can be found in the appendix.

The proposed AE effectively achieves task collaboration while keeping the inference cost almost unchanged. With the AE module, the tampered regions in features $F_{2}$ and $F_{3}$ can be enhanced and the false-positive noise can be reduced. Consequently, our AE module helps to extract better common features and thus benefits the generalist model.

\vspace{-0.2cm}
\subsection{Dynamic Weight Decoder}
\noindent \textbf{Key Idea.} Different types of tampered images result in a wide range of manipulation clues. For example, forged objects in natural images may have abnormal contrast or edge artifact~\cite{wang2022objectformer}, tampered text in document images might be visually consistent but has discontinuous BAG in frequency domain~\cite{CVPR2023DocTamper}, fake faces may have unnatural texture~\cite{hifi_net_xiaoguo}. These wide variations in tampering clues further cause a large variation of the encoded features of tampered regions. 
Merely using a fixed set of filters for the decoder causes it being confused by the diverse encoder features, especially in the unified training process. To this end, we propose to adaptively select the optimal decoder filters for each input image based on its characteristics. To achieve this, we propose the Dynamic Weight Decoder (DWD), as shown in the right of the Figure~\ref{fig: Modules}.

\noindent \textbf{Method.} In the proposed Dynamic Weight Decoder, the low-level input features are fused with high-level input features by Pyramid Pooling Module ~\cite{zhao2017pyramid} and Feature Pyramid Network~\cite{lin2017fpn} to obtain multi-scale features $F_{1}, F_{2}, F_{3}, F_{4}$. A global feature vector $V_{g}$ is obtained by average pooling $F_{1}$. The multi-scale features are channel dimension reduced and processed by a series of Dynamic Weight Filters (DWFs) with different dilation rates. The output features are channel dimension concatenated, reduced and are further processed by two DWFs to obtain the final prediction mask.

\noindent \textbf{Dynamic Weight Filters.} As shown in the top-right of Figure~\ref{fig: Modules}, to obtain the dynamic filters, we first average pool the input feature to obtain a current global representation $V_c$ (orange box), then interact $V_c$ with the global image vector $V_g$ (blue box) with a fully connected layer and identify the optimal dynamic filters $D_{opt}$ by weighted summation of four common convolutional filters. $A_{i}=\sigma(FC(V_c, V_g))$, $D_{opt}=
\sum_{i=1}^{4}A_i*Wi$, $\sigma$ is the sigmoid function, $FC$ is the linear layer, $W_i$ is the $i$th filter in the DWF.
Finally, we depth-wise convolve the input feature with $D_{opt}$ and then perform point-wise convolution with $1 \times 1$ conv-layer to obtain the output.

The proposed DWD is trained in an end-to-end manner, it achieves sample-wise filters selection by analyzing the characteristics of the input image, the input features and the forgery types in the initially predicted tampered region. The selected optimal filters effectively help the model to achieve generalism. 

\vspace{-0.2cm}
\subsection{Interpretation Module}
To provide more convincing and reliable forgery analysis by describing the image artifact in natural language, we introduce the Interpretation Module (IM), as shown on the right of Figure~\ref{fig: main}. 

Existing works directly input the tampered image into a multimodal LLM to describe the image artifact. However, due to the challenging nature of image forensics, the LLM often misidentifies the tampered region especially on multi-target and challenge scenarios (e.g. tampered documents). To address this issue, we propose to draw LLM's attention to the suspect regions by presenting it with the forgery localization mask predicted by the decoder. However, directly inputting the binary mask predicted by LM will lead to considerable ambiguity in dense text and dense face images, since adjacent instances may have very similar positions. 

To minimize the ambiguity and the difficulty for IM to understand the predicted tampered regions, we construct a visual reference prompt $I_{ref}$ by highlighting the mask predictions $I_{mask}$ (normed to 0-255) on the tampered image $I_{input}$ by pixel-wise addition: $I_{ref}=(I_{input}+I_{mask})/2$. We concatenate the $I_{ref}$ with $I_{input}$ along the longest side, and feed it into a multimodal LLM for text prediction. Besides clearly indicating the suspected region, our method can also minimize overfitting and forgetting for multimodal LLMs, since it does not change their original structure. 

\vspace{-0.2cm}
\section{Omni-273k Dataset}
To enable the Omni-IML for high-quality description of the tampered region, we construct the Omni-273k dataset, by querying GPT-4o to generate textual artifact descriptions for the tampered images in multiple major IML tasks.

\vspace{-0.2cm}
\subsection{Chain-of-Thoughts Automatic Annotation}
\noindent \textbf{Motivation.} Existing works generate textual descriptions for all tampered objects' content, position and artifact clues in a forged image by querying with a single prompt~\cite{xu2024fakeshield, huang2024sida}. These methods make progress on single-target and less challenging scenarios such as natural object images, but do not work well on tampered documents, scene texts and uncut deepfakes where image artifacts are less obvious and a tampered image usually contains multiple tampered instances. There are two main reasons for this: First, given the challenging nature of image forensics, accurately describing the content, location, and artifact of a tampered instance requires focused analysis. Simultaneous analysis of multiple targets leads to distractions, causing the model to confuse the artifact among different objects. Second, for challenging samples such as forged documents, the response of GPT-4o is often unconfident and partially incorrect. These problems prevent previous methods from being an adequate solution for unified image forgery analysis.

\begin{figure}[ht!]
  \centering
  \includegraphics[width=\linewidth]{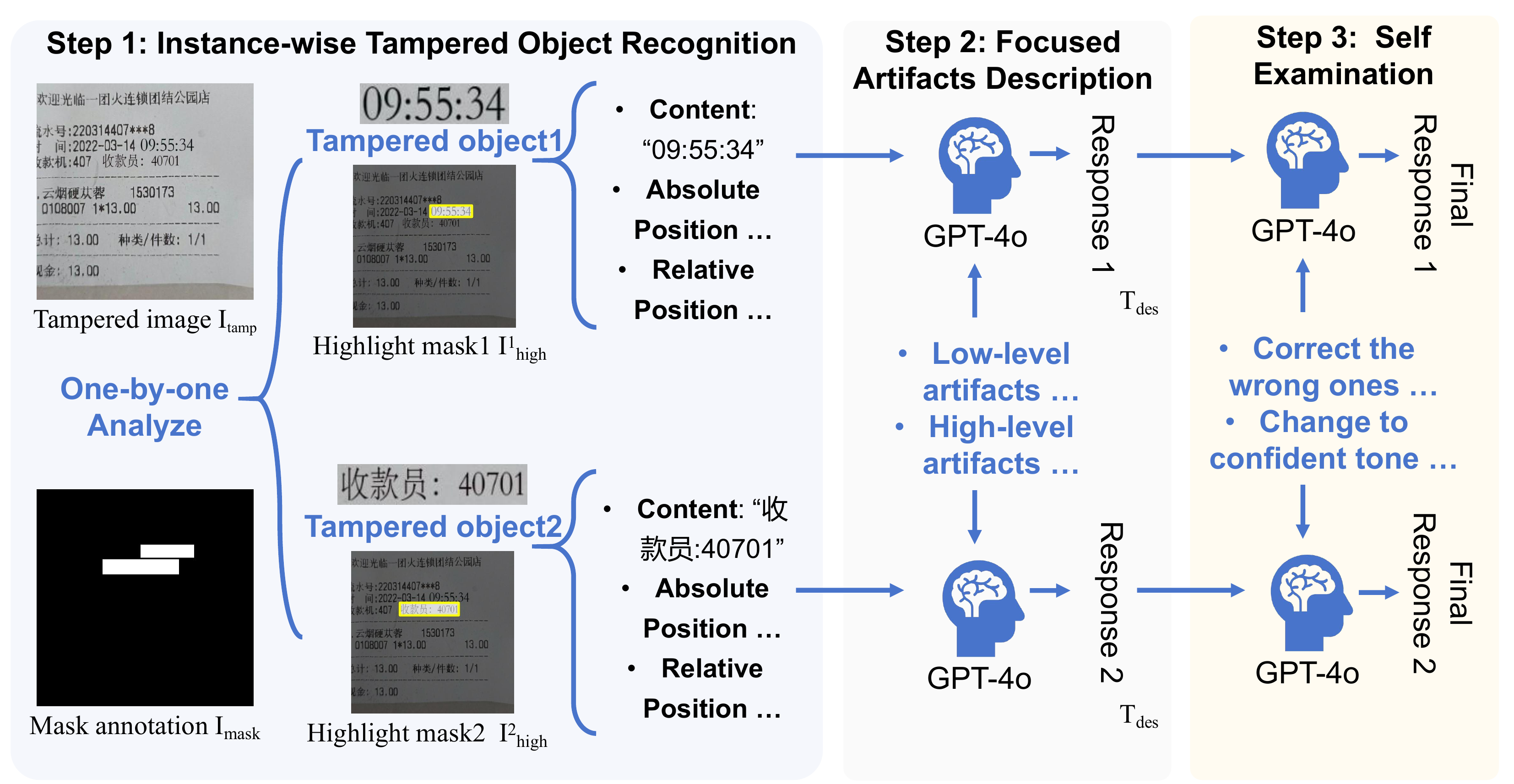}
  \vspace{-0.3cm}
  \caption{The proposed Chain-of-Thoughts Pipeline.}
  \label{fig: CoT}
  \vspace{-0.4cm}
\end{figure}

\noindent \textbf{Method.} To solve the above problems and improve annotation quality in all major scenarios. We propose a novel chain-of-thoughts pipeline, which consists of three steps, as shown in Figure~\ref{fig: CoT}:

\noindent \textbf{Step 1. Instance-wise Tampered Object Recognition.} In this step, we accurately recognize the content and position for each tampered region. Given the binary mask annotation $I_{mask}$ indicating the tampered region for a tampered image $I_{tamp}$, each of the connected components in $I_{mask}$ is a tampered instance. For text images, the OCR result for each tampered instance is the content, which is obtained through OCR engine. For other images, we highlight the $m$ tampered regions in $I_{tamp}$ respectively to get $m$ highlight masks $I^{n}_{high}$, $n\in[1,m]$. We prompt GPT-4o with $I_{tamp}$ and $I^{n}_{high}$ sequentially for each tampered instance to get its content (e.g. "A basketball", "The face of a man in blue shirts", more detailed in practice). Along with recognizing the content, we also obtain the position for each tampered instance. 

\begin{table*}[ht!]
\small
\centering
\setlength{\tabcolsep}{1pt}
\caption{Comparison on the interpretable IML datasets. 'CM' denotes copy-move, 'SP' denotes splicing, 'PT' denotes printing~\cite{CVPR2023DocTamper}, 'Forgery Num.' denotes the number of tampered images, 'Manual Num.' denotes the number of manually forged images.}
\vspace{-0.3cm}
\begin{tabular}{ccccccccccccccc}
\hline
Dataset &  & Year &  & Image Types &  & Tampering Methods &  & Target type &  & Forgery Num. &  & Manual Num. &  & Structured Format \\ \cline{1-1} \cline{3-3} \cline{5-5} \cline{7-7} \cline{9-9} \cline{11-11} \cline{13-13} \cline{15-15} 
MMTD-Set~\cite{xu2024fakeshield} &  & 2025 &  & Natural images, cut off faces. &  & CM, SP, AIGC &  & Single target &  & \textless 30, 000 &  & \textless 10, 000 &  & $\times$ \\
SID-Set~\cite{huang2024sida} &  & 2025 &  & Natural images. &  & AIGC &  & Single target &  & 200, 000 &  & 0 &  & $\times$ \\
\begin{tabular}[c]{@{}c@{}}Omni-273k\\ (Ours)\end{tabular} &  & 2025 &  & \textbf{\begin{tabular}[c]{@{}c@{}}Natural images, uncut faces,\\  various documents, scene text images.\end{tabular}} &  & \textbf{CM, SP, AIGC, PT} &  & \textbf{\begin{tabular}[c]{@{}c@{}}Singe target and\\ multiple targets\end{tabular}} &  & \textbf{273, 776} &  & \textbf{131, 658} &  & \textbf{\checkmark} \\ \hline
\end{tabular}
\label{tab: dataset}
\end{table*}

\noindent \textbf{Step 2. Focused Artifact Description.} In this step, we obtain detailed descriptions of image artifact for each tampered instance. For each tampered instance, we prompt GPT-4o with $I_{tamp}$, $I^{n}_{high}$, the previously recognized content and position, and an elaborate query (lists the common artifact perspectives) to obtain the detailed descriptions $T_{des}$ for low-level visual and high-level semantic artifacts. The full query is shown in the appendix. During this artifact description process, the annotator GPT-4o is \textbf{focused} in two aspects: First, the GPT-4o only focuses on one single tampered object, instead of all objects as in previous works. Second, except for the $I_{tamp}$, $I^{n}_{high}$, we also provide the previously recognized contents and positions. The GPT-4o can thus focus on the task of describing image artifact, rather than handling also multiple recognition tasks at the same time as in previous works. Focusing GPT-4o can effectively reduce hallucination and improve annotation quality. 

\noindent \textbf{Step 3. Self Examination.} In challenging samples where image artifacts are not obvious, the response of GPT-4o can be unconfident and incorrect, while manual filtering is costly. To this end, we propose to further improve the annotation quality by guiding GPT-4o to carefully examine and wash its previous response.
For each of the tampered object, we prompt GPT-4o with its previous response $T_{des}$, $I_{tamp}$, $I^{n}_{high}$, and the previously recognized contents and positions, then show it an example containing both an unconfident, incorrect answer and the corresponding manually corrected one (show in the appendix in detail). We have GPT-4o carefully examine and clean its response with the prompts and example. The final response has a significantly improved quality, is very closed to human annotation and is used as the final annotation. 

Our CoT pipeline differs from previous methods through \textbf{focused, step-by-step, self-examined} analysis that significantly improves annotation quality in unified IML analysis.

\smallskip

\subsection{Annotation Format} 
Different from previous works that use unstructured string as the annotation format of the artifact description, we adopt a structured JSON format. Our annotation for each tampered image is a list of $m$ items, where $m$ is the number of tampered instances in that image. Each of the $m$ items is a JSON dict with four key-value pairs: 

\noindent \textbf{(1)}. "Tampered Region", the value is the OCR of tampered text or the description of tampered object (e.g. 'A sleeping orange cat'). 

\noindent \textbf{(2)}. "Absolute Position", the value is the position of the tampered region relative to the entire image (e.g. 'Top left of the image'). 

\noindent \textbf{(3)}. "Relative Position", the value is the OCR of the text line containing the tampered text, or the position of the tampered object relative to other objects (e.g., "On the leftmost green table").

\noindent \textbf{(4)}. "Artifacts", the value is the artifact description of the tampered region. It is also a JSON dict, its keys are the titles for different artifacts (e.g. "Textural Artifacts", "Semantic Artifacts") and its values are the detailed descriptions under the titles. 

Compared to unstructured string format that can only support coarse model evaluation on the entire output~\cite{xu2024fakeshield, huang2024sida}, our \textbf{novel} JSON format \textbf{enables much finer-grained and in-depth analysis} based on detailed items (e.g. content, positions or artifacts). The dataset samples are present in the appendix.

\subsection{Dataset Construction}
We apply our propose Chain-of-Thought pipeline on tampered images from the most commonly used datasets of the four major IML tasks, including: AutoSplice~\cite{autosplice}, CASIAv1\&v2~\cite{casia}, Coverage~\cite{coverage}, CocoGlide~\cite{guillaro2023trufor}, IMD20~\cite{imd20}, NIST16~\cite{nist16}, MIML~\cite{Qu_2024_CVPR} of natural IML, DocTamper~\cite{CVPR2023DocTamper}, SACP~\cite{sacp} of document IML, OpenForensics~\cite{le2021openforensics} of face IML and Tampered-IC13~\cite{wang2022tic}, OSTF~\cite{ostf} of scene text IML.A very minority of samples with bad annotations are filtered out. Finally, there are a total of 273, 776 samples in our dataset. 217, 907 images from CASIAv2, MIML, and the training sets of DocTamper, SACP, OpenForensics, Tampered-IC13 and OSFT are split as the training set of Omni-273k, the rest 55, 869 images are split as the test set.  More statistics can be found in the appendix. 

\subsection{Dataset Highlights}
As shown in Table~\ref{tab: dataset}, our Omni-273k is is \textbf{the first} interpretable IML dataset to cover all the four major image types of IML, significantly outperforming existing datasets in multiple perspectives:

\noindent $\bullet$ \textbf{Large-Scale:} Our dataset has the largest scale and even 10+ times more manually tampered (high quality) images than others.

\noindent  $\bullet$ \textbf{Comprehensive:} Our datatset includes all the four major image types, all the common tampering methods, and contains both single and multi-target (>1 tampered objects in an image) samples.

\noindent $\bullet$ \textbf{High-Quality:} Our dataset is \textbf{the first} to include structured annotation format to enable fine-grained analysis. Our dataset is constructed and cleaned by our CoT pipeline to ensure quality.

We believe that our large-scale, comprehensive and high-quality dataset can considerably benefit the community.

\begin{table*}[ht!]
\small
\setlength{\tabcolsep}{2.3pt}
\caption{Comparison study on natural image manipulation localization.}
\vspace{-0.2cm}
\begin{tabular}{ccccccccccccccccccccccccccccccccc}
\hline
\multirow{2}{*}{Method} &  & \multicolumn{3}{c}{CASIAv1~\cite{casia}} &  & \multicolumn{3}{c}{Coverage~\cite{coverage}} &  & \multicolumn{3}{c}{NIST16~\cite{nist16}} &  & \multicolumn{3}{c}{IMD20~\cite{imd20}} &  & \multicolumn{3}{c}{CocoGlide~\cite{guillaro2023trufor}} &  & \multicolumn{3}{c}{AutoSplice~\cite{autosplice}} &  & \multicolumn{3}{c}{Avg. (w.o. IMD20)} &  & \multicolumn{3}{c}{Avg. (w/ IMD20)} \\ \cline{3-5} \cline{7-9} \cline{11-13} \cline{15-17} \cline{19-21} \cline{23-25}  \cline{27-29} \cline{31-33} 
 &  & IoU &  & F1 &  & IoU &  & F1 &  & IoU &  & F1 &  & IoU &  & F1 &  & IoU &  & F1 &  & IoU &  & F1 &  & IoU &  & F1 &  & IoU &  & F1 \\ \cline{1-1} \cline{3-3} \cline{5-5} \cline{7-7} \cline{9-9} \cline{11-11} \cline{13-13} \cline{15-15} \cline{17-17} \cline{19-19} \cline{21-21} \cline{23-23} \cline{25-25}  \cline{27-27} \cline{29-29} \cline{31-31} \cline{33-33} 
ManTraNet~\cite{wu2019mantra} &  & .086 &  & .130 &  & .181 &  & .271 &  & .040 &  & .062 &  & .098 &  & .146 &  & .155 &  & .203 &  & .120 &  & .179  &  & .116 &  & .169 &  & .113 &  & .165 \\
RRU-Net~\cite{bi2019rru} &  & .330 &  & .380 &  & .165 &  & .260 &  & .080 &  & .129 &  & .169 &  & .256 &  & .223 &  & .304 &  & .271 &  & .387  &  & .214 &  & .292 &  & .206 &  & .286 \\
MVSS-Net~\cite{dong2022mvss} &  & .403 &  & .435 &  & .389 &  & .454 &  & .243 &  & .294 &  & .243 &  & .294 &  & .276 &  & .357 &  & .241 &  & .334  &  & .310 &  & .375 &  & .299 &  & .361 \\
PSCC-Net~\cite{liu2022pscc} &  & .410 &  & .463 &  & .340 &  & .446 &  & .067 &  & .110 &  & .115 &  & .192 &  & .333 &  & .422 &  & .447 &  & .558  &  & .319 &  & .400 &  & .285 &  & .365 \\
CAT-Net~\cite{kwon2022learning} &  & .684 &  & .738 &  & .238 &  & .292 &  & .238 &  & .302 &  & - &  & - &  & .290 &  & .366 &  & \textbf{.579} &  & \textbf{.676}  &  & .406 &  & .475 &  & - &  & - \\
IF-OSN~\cite{wu2022robust} &  & .465 &  & .509 &  & .181 &  & .268 &  & .247 &  & .326 &  & .259 &  & .364 &  & .207 &  & .264 &  & .395 &  & .509  &  & .299 &  & .375 &  & .292 &  & .373 \\
EVP~\cite{liu2023explicit} &  & .438 &  & .502 &  & .078 &  & .114 &  & .188 &  & .239 &  & .177 &  & .268 &  & .084 &  & .118 &  & .136 &  & .203  &  & .185 &  & .235 &  & .184 &  & .241 \\
TruFor~\cite{guillaro2023trufor} &  & .630 &  & .692 &  & .446 &  & .522 &  & .279 &  & .348 &  & - &  & - &  & .294 &  & .362 &  & .393 &  & .503  &  & .408 &  & .485 &  & - &  & - \\
APSC-Net~\cite{Qu_2024_CVPR} &  & \textbf{.810} &  & \textbf{.848} &  & .498 &  & .568 &  & .525 &  & .590 &  & \textbf{.679} &  & \textbf{.760} &  & .392 &  & .455 &  & .409 &  & .508  &  & .526 &  & .594 &  & .552 &  & .621 \\
Omni-IML (Ours) &  & .796 &  & .832 &  & \textbf{.515} &  & \textbf{.590} &  & \textbf{.522} &  & \textbf{.597} &  & .662 &  & .742 &  & \textbf{.510} &  & \textbf{.575} &  & .454 &  & .550  &  & \textbf{.559} &  & \textbf{.629} &  & \textbf{.577} &  & \textbf{.648} \\ \hline
\end{tabular}
\vspace{-0.1cm}
\label{tab: natcomp}
\end{table*}

\begin{table}[ht!]
\small
\centering
\setlength{\tabcolsep}{1pt}
\caption{Comparison on document IML.}
\vspace{-0.2cm}
% \usepackage{multirow}
\begin{tabular}{ccccccccccccccccc}
\hline
\multirow{2}{*}{Method} &  & \multicolumn{3}{c}{SACP} &  & \multicolumn{3}{c}{DocTamper-Test} &  & \multicolumn{3}{c}{DocTamper-FCD} &  & \multicolumn{3}{c}{DocTamper-SCD} \\ \cline{3-5} \cline{7-9} \cline{11-13} \cline{15-17} 
 &  & IoU &  & F1 &  & IoU &  & F1 &  & IoU &  & F1 &  & IoU &  & F1 \\ \cline{1-1} \cline{3-3} \cline{5-5} \cline{7-7} \cline{9-9} \cline{11-11} \cline{13-13} \cline{15-15} \cline{17-17} 
DFCN &  & .466 &  & .607 &  & - &  & - &  & - &  & - &  & - &  & - \\
MVSS-Net &  & .401 &  & .534 &  & - &  & - &  & - &  & - &  & - &  & - \\
SE-Net &  & .459 &  & .587 &  & - &  & - &  & - &  & - &  & - &  & - \\
RRU-Net &  & .517 &  & .651 &  & - &  & - &  & - &  & - &  & - &  & - \\
CFL-Net &  & .433 &  & .571 &  & - &  & - &  & - &  & - &  & - &  & - \\
TIFDM &  & .576 &  & .703 &  & - &  & - &  & - &  & - &  & - &  & - \\
MantraNet &  & - &  & - &  & .180 &  & .123 &  & .170 &  & .209 &  & .160 &  & .157 \\
MVSS-Net &  & - &  & - &  & .430 &  & .431 &  & .410 &  & .424 &  & .400 &  & .414 \\
PSCC-Net &  & - &  & - &  & .170 &  & .384 &  & .160 &  & .420 &  & .190 &  & .374 \\
BEiT-UPer &  & - &  & - &  & .590 &  & .501 &  & .350 &  & .487 &  & .340 &  & .402 \\
Swin-UPer &  & - &  & - &  & .700 &  & .638 &  & .410 &  & .546 &  & .510 &  & .574 \\
CAT-Netv2 &  & - &  & - &  & .710 &  & .721 &  & .600 &  & .741 &  & .540 &  & .670 \\
DTD &  & - &  & - &  & .828 &  & .792 &  & .749 &  & .816 &  & \textbf{.691} &  & \textbf{.754} \\
Ours &  & \textbf{.659} &  & \textbf{.778} &  & \textbf{.856} &  & \textbf{.806} &  & \textbf{.851} &  & \textbf{.894} &  & .664 &  & .740 \\ \hline
\end{tabular}
\vspace{-0.4cm}
\label{tab: doccomp}
\end{table}

\begin{table}[ht!]
\small
\vspace{+0.3cm}
\caption{Comparison on face (left) and scene text (right) IML.}
\vspace{-0.2cm}
\setlength{\tabcolsep}{1.25pt}
\begin{tabular}{ccccccccccccccc}
\cline{1-5} \cline{7-15}
\multicolumn{5}{c}{Face IML} & \multirow{10}{*}{} & \multicolumn{9}{c}{Scene Text IML} \\ \cline{1-5} \cline{7-15} 
Method &  & IoU &  & F1 &  & \multirow{2}{*}{Method} &  & \multicolumn{3}{c}{T-IC13} &  & \multicolumn{3}{c}{OSTF} \\ \cline{1-1} \cline{3-3} \cline{5-5} \cline{9-11} \cline{13-15} 
ManTra-Net &  & .720 &  & .837 &  &  &  & IoU &  & F1 &  & IoU &  & F1 \\ \cline{7-7} \cline{9-9} \cline{11-11} \cline{13-13} \cline{15-15} 
HPFCN &  & .726 &  & .841 &  & DeepLabV3+ &  & .722 &  & .837 &  & .290 &  & .442 \\
MVSS-Net &  & .701 &  & .824 &  & HRNetv2 &  & .731 &  & .845 &  & .295 &  & .444 \\
CAT-Net &  & .832 &  & .908 &  & BEiT-UPer &  & .709 &  & .830 &  & .276 &  & .437 \\
DOAGAN &  & .732 &  & .845 &  & SegFormer &  & \textbf{.778} &  & \textbf{.875} &  & .302 &  & .450 \\
HiFiNet &  & .749 &  & .856 &  & Swin-UPer &  & .773 &  & .872 &  & .307 &  & .452 \\
MoNFAP &  & .902 &  & .949 &  & UPOCR &  & .716 &  & .835 &  & .281 &  & .439 \\
Omni-IML (Ours) &  & \textbf{.923} &  & \textbf{.957} &  & Omni-IML (Ours) &  & .741 &  & .851 &  & \textbf{.456} &  & \textbf{.627} \\ \cline{1-5} \cline{7-15} 
\label{tab: facesce}
\vspace{-0.6cm}
\end{tabular}
\end{table}

\begin{figure}[hb!]
\vspace{-0.3cm}
  \centering
  \includegraphics[height=50pt]{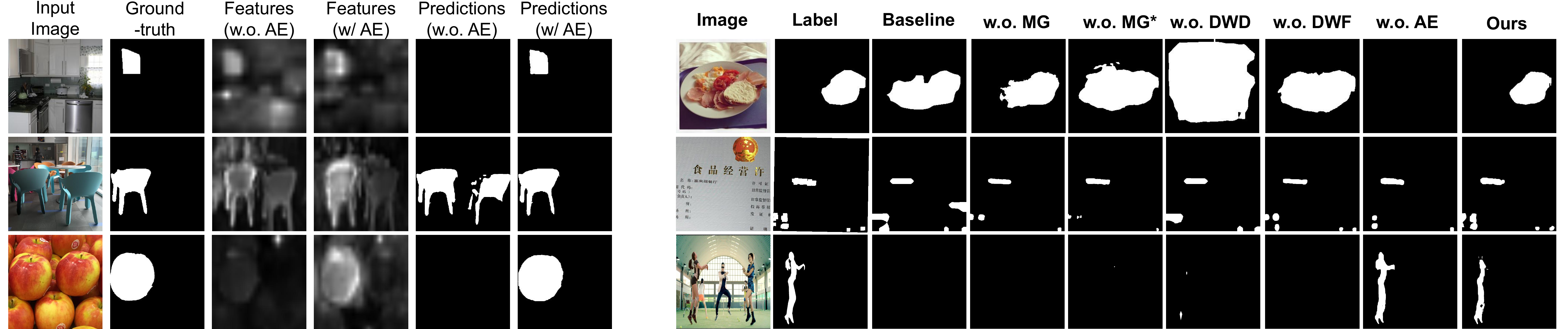}
  \vspace{-0.2cm}
  \caption{The proposed Chain-of-Thoughts Pipeline.}
  \label{fig: viz}
  \vspace{-0.4cm}
\end{figure}

\section{Experiments}
\subsection{Experiment Setup}
The training data for both our localization and interpretation modules include four parts: (1) Natural style images, including tampCOCO~\cite{kwon2022learning}, CASIAv2~\cite{casia}, MIML~\cite{Qu_2024_CVPR}, and COCO~\cite{lin2014microsoft} datasets.
(2) Document images, including the training sets of SACP~\cite{sacp} and DocTamper~\cite{CVPR2023DocTamper}.
(3) Face images, using the OpenForensics~\cite{le2021openforensics} training set and (4) Scene text images, using the Tampered-IC13~\cite{wang2022tic} training set.
The test data for both modules also includes four parts: CASIAv1~\cite{casia}, Coverage~\cite{coverage}, NIST16~\cite{nist16}, IMD20~\cite{imd20}, CocoGlide~\cite{guillaro2023trufor} and AutoSplice~\cite{autosplice} for natural images, the test sets of SACP~\cite{sacp} and DocTamper for document images, the test sets of OpenForensics and OSTF~\cite{ostf} for face and scene text images respectively. Such comprehensive benchmarks can thoroughly evaluate both in-domain and out-of-domain model performance on various image types and tampering methods (both traditional and AIGC).

\noindent \textbf{Localization Module}. The backbone model of our Omni-IML is ConvNeXt-Base~\cite{liu2022convnet} initialized with its official ADE20k~\cite{ade20k} pre-trained weights, following previous works~\cite{Yu_2024_CVPR, Qu_2024_CVPR}. The Omni-IML is trained with the cross-entropy loss for 400k iterations, using the AdamW optimizer~\cite{adamw}, with a batch size of 16 and an input size of $512\times512$. The initial learning rate is set to 1e-4 and decays to 1e-6 in a linear schedule. A fixed threshold of 0.5 is used to binarize model predictions during inference. Pixel-level Intersection over Union (IoU) and binary F1~\cite{ma2024benco} are used to evaluate model performance.

\noindent \textbf{Interpretation Module}. The multimodal LLM in our interpretation module is LoRA~\cite{hu2022lora} fine-tuned 60k iterations with a batch-size of 16, a LoRA rank of 64, an AdamW optimizer with learning rate decayed from 1e-4 to 0.

\subsection{Comparison Study on Forgery Localization}
The proposed Omni-IML is evaluated on all of the four major IML tasks using a single set of model parameters, without any task-specific or benchmark-wise fine-tuning. The comparison results on natural IML, document IML, face IML and scene text IML are shown in Table~\ref{tab: natcomp}, \ref{tab: doccomp}, \ref{tab: facesce} respectively. Evidently, our generalist Omni-IML can simultaneously outperform existing specialized methods on each individual task, demonstrating the strong generalization ability. This is because our Omni-IML can adaptively select the optimal input modality and decoder parameters for each sample, effectively producing the best features for IML on different image types. In addition, the Anomaly Enhancement module suppresses feature noise and reduces model confusion in joint training, which further improves our generalist model.

To further explore the generalist capability of previous IML methods, we re-train the state-of-the-art models with their official model code, the same training data and pipeline as ours, the results are shown in Table~\ref{tab: jointtrain}. In Table~\ref{tab: jointtrain}, the left part is the performance of the models trained on specific IML tasks. Evidently, all the models perform well on only one task. For example, the TruFor trained on one task (Natural IML) has an IoU score of 0.485 on Natural IML task but its IoU score<0.1 on other IML tasks. The right part of Table~\ref{tab: jointtrain} is the performance of models jointly trained on all tasks. The previous IML methods suffer more performance degradation and show much worse average performance than our Omni-IML in joint training. For example, for TIFDM, its IoU on document IML is 0.568 when it's trained on document IML task only, but its IoU on document IML is 0.499 when it's trained on all the four IML tasks, which is a degradation of 6.9 points. While for our Omni-IML, the IoU degradation brought by joint training 
 is merely 1.6 points ($0.774-0.758$) on document IML task. This is because the previous methods rely heavily on designs and strategies targeted at one IML task, and such designs and strategies usually do not work as well on other image types (e.g. noise filters, edge enhancement and object-level attention are beneficial for natural images but not for document images). Moreover, the tampering features among diverse image types differ a lot from each other. Without an sample-adaptive and noise-suppression design, the previous methods are challenged to simultaneously learn them well. In contrast, our Omni-IML does not rely on task-dependent design and benefits from the adaptive selection of optimal encoding modality, decoder parameters, and noise-suppression. Consequently, our Omni-IML demonstrates strong generalization across image types and has minimal performance degradation during joint training. Qualitative results for visual comparison are presented in Figure~\ref{fig: viz}.

\begin{table}[t!]
\small
\vspace{+0.3cm}
\caption{Comparison study of models trained on a specific task and all tasks, using the pixel-level IoU metric. 'Nat.' denotes natural IML, 'Doc.' denotes document IML, 'S.T.' denotes scene text IML, 'Face' denotes face IML.}
\vspace{-0.2cm}
\setlength{\tabcolsep}{1.25pt}
\begin{tabular}{ccccccccccccccccccc}
\hline
\multirow{2}{*}{Method} &  & \multicolumn{7}{c}{Trained on one task} &  & \multicolumn{9}{c}{Trained on all tasks} \\ \cline{3-9} \cline{11-19} 
 &  & Nat. &  & Doc. &  & S.T. &  & Face &  & Nat. &  & Doc. &  & S.T. &  & Face &  & Avg. \\ \cline{1-1} \cline{3-3} \cline{5-5} \cline{7-7} \cline{9-9} \cline{11-11} \cline{13-13} \cline{15-15} \cline{17-17} \cline{19-19} 
CAT-Net (Nat.) &  & .462 &  & .089 &  & .097 &  & .112 &  & .437 &  & .571 &  & .530 &  & .874 &  & .603 \\
TruFor (Nat.) &  & .485 &  & .054 &  & .079 &  & .098 &  & .453 &  & .555 &  & .542 &  & .898 &  & .612 \\
APSC-Net (Nat.) &  & .552 &  & .047 &  & .067 &  & .076 &  & .511 &  & .607 &  & .558 &  & .901 &  & .644 \\
TIFDM (Doc.) &  & .020 &  & .568 &  & .034 &  & .001 &  & .382 &  & .499 &  & .489 &  & .834 &  & .551 \\
DTD (Doc.) &  & .043 &  & .605 &  & .069 &  & .002 &  & .396 &  & .522 &  & .481 &  & .822 &  & .555 \\
Swin-UPer (S.T.) &  & .045 &  & .033 &  & .586 &  & .037 &  & .425 &  & .547 &  & .525 &  & .850 &  & .587 \\
MoNFAP (Face) &  & .090 &  & .022 &  & .004 &  & .902 &  & .398 &  & .534 &  & .499 &  & .845 &  & .569 \\ \hline
Ours (Nat.) &  & \textbf{.595} &  & .063 &  & .092 &  & .104 &  & - &  & - &  & - &  & - &  & - \\
Ours (Doc.) &  & .061 &  & \textbf{.774} &  & .087 &  & .002 &  & - &  & - &  & - &  & - &  & - \\
Ours (S.T.) &  & .087 &  & .079 &  & \textbf{.623} &  & .030 &  & - &  & - &  & - &  & - &  & - \\
Ours (Face) &  & .058 &  & .006 &  & .008 &  & \textbf{.932} &  & - &  & - &  & - &  & - &  & - \\
Ours (All) &  & - &  & - &  & - &  & - &  & \textbf{.577} &  & \textbf{.758} &  & \textbf{.599} &  & \textbf{.923} &  & \textbf{.714} \\ \hline
\end{tabular}
\label{tab: jointtrain}
\end{table}

\begin{table}[t!]
\small
\vspace{+0.3cm}
\caption{Ablation study on the proposed modules.}
\vspace{-0.2cm}
\setlength{\tabcolsep}{1.25pt}
\begin{tabular}{ccccccccccccccccccccc}
\hline
 &  & \multicolumn{3}{c}{Natural} &  & \multicolumn{3}{c}{Document} &  & \multicolumn{3}{c}{Scene Text} &  & \multicolumn{3}{c}{Face} &  & \multicolumn{3}{c}{Average} \\ \cline{3-5} \cline{7-9} \cline{11-13} \cline{15-17} \cline{19-21}
\multirow{-2}{*}{Ablation} &  & IoU &  & F1 &  & IoU &  & F1 &  & IoU &  & F1 &  & IoU &  & F1 &  & IoU &  & F1 \\ \cline{1-1} \cline{3-3} \cline{5-5} \cline{7-7} \cline{9-9} \cline{11-11} \cline{13-13} \cline{15-15} \cline{17-17} \cline{19-19} \cline{21-21} 
{Baseline} &  & {.432} &  & .567 &  & .516 &  & .649 &  & .492 &  & .670 &  & .848 &  & .912 &  & .572 &  & .699 \\
w.o. MG &  & .489 &  & .601 &  & .615 &  & .725 &  & .537 &  & .696 &  & .875 &  & .925 &  & .629 &  & .737 \\
w.o. MG* &  & .538 &  & .625 &  & .634 &  & .742 &  & .561 &  & .712 &  & .904 &  & .947 &  & .659 &  & .756 \\
w.o. DWD &  & .463 &  & .582 &  & .589 &  & .703 &  & .515 &  & .683 &  & .867 &  & .922 &  & .609 &  & .722 \\
w.o. DW &  & .542 &  & .628 &  & .728 &  & .788 &  & .568 &  & .717 &  & .912 &  & .951 &  & .688 &  & .771 \\
w.o. AE &  & .548 &  & .632 &  & .707 &  & .776 &  & .570 &  & .718 &  & .916 &  & .953 &  & .685 &  & .769 \\
Ours &  & {\textbf{.577}} & \textbf{} & \textbf{.648} & \textbf{} & \textbf{.758} & \textbf{} & \textbf{.805} &  & \textbf{.599} & \textbf{} & \textbf{.739} & \textbf{} & \textbf{.923} & \textbf{} & \textbf{.957} &  & \textbf{.714} &  & \textbf{.787} \\ \hline
\end{tabular}
\label{tab: ablmain}
\end{table}

\begin{table*}[ht!]
\small
\setlength{\tabcolsep}{0.8pt}
\caption{Fine-grained study of interpretation ability. 'Text Rec.' denotes tampered text recognition, 'Abs. Pos' denotes absolute position, 'Rel. Pos.' denotes relative position, 'Obj. Rec.' denotes Tampered object recognition. 'OCR' denotes the OCR accuracy metric. 'Acc' denotes accuracy. 'MRB' denotes the mean score of ROUGE-L and BLEU}
\vspace{-0.3cm}
\begin{tabular}{ccccccccccccccccccccccccccccccccccccccccccccccccccccccccc}
\hline
\multirow{4}{*}{Method} &  & \multicolumn{11}{c}{Document Image} &  & \multicolumn{11}{c}{Scene Text Image} &  & \multicolumn{15}{c}{Face Image} &  & \multicolumn{13}{c}{Natural Image} &  & \multirow{4}{*}{\begin{tabular}[c]{@{}c@{}}Average\\ Score\end{tabular}} \\ \cline{3-13} \cline{15-25} \cline{27-41} \cline{43-55}
 &  & \multirow{2}{*}{\begin{tabular}[c]{@{}c@{}}Text.\\ Rec.\end{tabular}} &  & \multirow{2}{*}{\begin{tabular}[c]{@{}c@{}}Abs.\\ Pos.\end{tabular}} &  & \multirow{2}{*}{\begin{tabular}[c]{@{}c@{}}Rel.\\ Pos.\end{tabular}} &  & \multicolumn{5}{c}{\multirow{2}{*}{\begin{tabular}[c]{@{}c@{}}artifact\\ Description\end{tabular}}} &  & \multirow{2}{*}{\begin{tabular}[c]{@{}c@{}}Text.\\ Rec.\end{tabular}} & \multirow{2}{*}{} & \multirow{2}{*}{\begin{tabular}[c]{@{}c@{}}Abs.\\ Pos.\end{tabular}} &  & \multirow{2}{*}{\begin{tabular}[c]{@{}c@{}}Rel.\\ Pos.\end{tabular}} &  & \multicolumn{5}{c}{\multirow{2}{*}{\begin{tabular}[c]{@{}c@{}}artifact\\ Description\end{tabular}}} &  & \multicolumn{3}{c}{\multirow{2}{*}{\begin{tabular}[c]{@{}c@{}}Obj.\\ Rec.\end{tabular}}} &  & \multirow{2}{*}{\begin{tabular}[c]{@{}c@{}}Abs.\\ Pos.\end{tabular}} &  & \multicolumn{3}{c}{\multirow{2}{*}{\begin{tabular}[c]{@{}c@{}}Rel.\\ Pos.\end{tabular}}} &  & \multicolumn{5}{c}{\multirow{2}{*}{\begin{tabular}[c]{@{}c@{}}artifact\\ Description\end{tabular}}} &  & \multicolumn{3}{c}{\multirow{2}{*}{\begin{tabular}[c]{@{}c@{}}Obj.\\ Rec.\end{tabular}}} &  & \multirow{2}{*}{\begin{tabular}[c]{@{}c@{}}Abs.\\ Pos.\end{tabular}} &  & \multicolumn{3}{c}{\multirow{2}{*}{\begin{tabular}[c]{@{}c@{}}Rel.\\ Pos.\end{tabular}}} &  & \multicolumn{3}{c}{\multirow{2}{*}{\begin{tabular}[c]{@{}c@{}}artifact\\ Description\end{tabular}}} &  &  \\
 &  &  &  &  &  &  &  & \multicolumn{5}{c}{} &  &  &  &  &  &  &  & \multicolumn{5}{c}{} &  & \multicolumn{3}{c}{} &  &  &  & \multicolumn{3}{c}{} &  & \multicolumn{5}{c}{} &  & \multicolumn{3}{c}{} &  &  &  & \multicolumn{3}{c}{} &  & \multicolumn{3}{c}{} &  &  \\ \cline{3-3} \cline{5-5} \cline{7-7} \cline{9-13} \cline{15-15} \cline{17-17} \cline{19-19} \cline{21-25} \cline{27-29} \cline{31-31} \cline{33-35} \cline{37-41} \cline{43-45} \cline{47-47} \cline{49-51} \cline{53-55}
 &  & OCR &  & Acc &  & OCR &  & Acc &  & \multicolumn{3}{c}{MRB} &  & OCR &  & Acc &  & OCR &  & Acc &  & \multicolumn{3}{c}{MRB} &  & \multicolumn{3}{c}{MRB} &  & Acc &  & \multicolumn{3}{c}{MRB} &  & Acc &  & \multicolumn{3}{c}{MRB} &  & \multicolumn{3}{c}{MRB} &  & Acc &  & \multicolumn{3}{c}{MRB} &  & Acc &  & MRB &  &  \\ \hline
\multicolumn{57}{c}{Supervised Fine-Tuned Models} \\ \hline
DeepSeek-VL 7B~\cite{lu2024deepseek} &  & .346 &  & .426 &  & .461 &  & .519 &  & \multicolumn{3}{c}{.186} &  & .440 &  & .496 &  & .512 &  & .552 &  & \multicolumn{3}{c}{.180} &  & \multicolumn{3}{c}{.792} &  & .926 &  & \multicolumn{3}{c}{.626} &  & .851 &  & \multicolumn{3}{c}{.344} &  & \multicolumn{3}{c}{.223} &  & .702 &  & \multicolumn{3}{c}{.279} &  & \textbf{.806} &  & .272 &  & .497 \\
MiniCPM-V-2.6 8B~\cite{yao2024minicpm} &  & .229 &  & .373 &  & .384 &  & .478 &  & \multicolumn{3}{c}{.173} &  & .492 &  & .464 &  & .546 &  & .578 &  & \multicolumn{3}{c}{.194} &  & \multicolumn{3}{c}{.758} &  & .898 &  & \multicolumn{3}{c}{.583} &  & .831 &  & \multicolumn{3}{c}{.330} &  & \multicolumn{3}{c}{.199} &  & .669 &  & \multicolumn{3}{c}{.251} &  & .790 &  & .268 &  & .474 \\
Intern2-VL 2B~\cite{chen2024internvl} &  & .276 &  & .379 &  & .389 &  & .480 &  & \multicolumn{3}{c}{.172} &  & .421 &  & .449 &  & .473 &  & .530 &  & \multicolumn{3}{c}{.178} &  & \multicolumn{3}{c}{.757} &  & .896 &  & \multicolumn{3}{c}{.591} &  & .828 &  & \multicolumn{3}{c}{.331} &  & \multicolumn{3}{c}{.197} &  & .643 &  & \multicolumn{3}{c}{.239} &  & .778 &  & .266 &  & .464 \\
Qwen2.5-VL 3B~\cite{bai2025qwen2} &  & .292 &  & .348 &  & .419 &  & .539 &  & \multicolumn{3}{c}{.197} &  & .489 &  & .451 &  & .553 &  & .572 &  & \multicolumn{3}{c}{.202} &  & \multicolumn{3}{c}{.738} &  & .801 &  & \multicolumn{3}{c}{.599} &  & .775 &  & \multicolumn{3}{c}{.340} &  & \multicolumn{3}{c}{.241} &  & .694 &  & \multicolumn{3}{c}{.310} &  & .771 &  & .298 &  & .481 \\
Qwen2.5-VL 7B~\cite{bai2025qwen2} &  & .312 &  & .381 &  & .429 &  & .521 &  & \multicolumn{3}{c}{.202} &  & .580 &  & .536 &  & .640 &  & .614 &  & \multicolumn{3}{c}{.217} &  & \multicolumn{3}{c}{.768} &  & .879 &  & \multicolumn{3}{c}{.630} &  & .794 &  & \multicolumn{3}{c}{.342} &  & \multicolumn{3}{c}{.270} &  & \textbf{.752} &  & \multicolumn{3}{c}{\textbf{.342}} &  & .749 &  & .301 &  & .512 \\ \hline
\multicolumn{57}{c}{Supervised Fine-Tuned Models with our proposed method (the reference visual prompt in interpretation module)} \\ \hline
Ours + Intern2-VL 2B &  & .603 &  & .586 &  & .647 &  & .638 &  & \multicolumn{3}{c}{.230} &  & .610 &  & .582 &  & .635 &  & .690 &  & \multicolumn{3}{c}{.224} &  & \multicolumn{3}{c}{.769} &  & .938 &  & \multicolumn{3}{c}{.587} &  & \textbf{.847} &  & \multicolumn{3}{c}{.339} &  & \multicolumn{3}{c}{.233} &  & .730 &  & \multicolumn{3}{c}{.281} &  & .781 &  & .265 &  & .561 (\textbf{+21\%}) \\
Ours + Qwen2.5-VL 3B &  & .640 &  & .555 &  & .690 &  & .667 &  & \multicolumn{3}{c}{.244} &  & \textbf{.719} &  & \textbf{.614} &  & \textbf{.766} &  & .678 &  & \multicolumn{3}{c}{.239} &  & \multicolumn{3}{c}{.815} &  & .940 &  & \multicolumn{3}{c}{.668} &  & .843 &  & \multicolumn{3}{c}{.366} &  & \multicolumn{3}{c}{\textbf{.272}} &  & .748 &  & \multicolumn{3}{c}{.330} &  & .748 &  & \textbf{.302} &  & .592 (\textbf{+23\%}) \\
Ours + Qwen2.5-VL 7B &  & \textbf{.647} &  & \textbf{.569} &  & \textbf{.692} &  & \textbf{.681} &  & \multicolumn{3}{c}{\textbf{.252}} &  & .712 &  & .609 &  & .735 &  & \textbf{.742} &  & \multicolumn{3}{c}{\textbf{.250}} &  & \multicolumn{3}{c}{\textbf{.817}} &  & \textbf{.946} &  & \multicolumn{3}{c}{\textbf{.668}} &  & .862 &  & \multicolumn{3}{c}{\textbf{.380}} &  & \multicolumn{3}{c}{.256} &  & .743 &  & \multicolumn{3}{c}{.314} &  & .765 &  & .296 &  & \textbf{.597 (+16\%}) \\
\hline
\end{tabular}
\label{tab: mllm2}
\end{table*}

\begin{table}[t!]
\small
\vspace{+0.3cm}
\caption{Experiments on the interpretation ability. 'Cos. Sim.' denotes the cosine similarity of the paragraph vectors.}
\vspace{-0.2cm}
\setlength{\tabcolsep}{1pt}
\begin{tabular}{ccccccccccc}
\hline
Method &  & Cos. Sim. &  & ROUGE-2 &  & ROUGE-L &  & BLEU &  & Average \\ \hline
\multicolumn{9}{c}{Pre-trained models (\textbf{zero-shot}).} \\ \hline
GPT-4o &  & .975 &  & .112 &  & .201 &  & .074 &  & .341  \\
Intern2-VL 2B~\cite{chen2024internvl} &  & .973 &  & .063 &  & .135 &  & .017 &  & .297  \\
Qwen2.5-VL 3B~\cite{bai2025qwen2} &  & .964 &  & .077 &  & .151 &  & .020 &  & .303  \\
Qwen2.5-VL 7B~\cite{bai2025qwen2} &  & .972 &  & .094 &  & .175 &  & .035 &  & .319  \\ \hline
\multicolumn{9}{c}{Supervise Fine-Tuned (\textbf{SFT}) models.} \\ \hline
DeepSeek-VL 7B~\cite{lu2024deepseek} &  & .993 &  & .385 &  & .441 &  & .308 &  & .532  \\
MiniCPM-V-2.6 8B~\cite{yao2024minicpm} &  & .992 &  & .378 &  & .439 &  & .287 &  & .524  \\
Intern2-VL 2B~\cite{chen2024internvl} &  & .993 &  & .372 &  & .432 &  & .293 &  & .523  \\
Qwen2.5-VL 3B~\cite{bai2025qwen2} &  & .994 &  & .403 &  & .456 &  & .316 &  & .542  \\
Qwen2.5-VL 7B~\cite{bai2025qwen2} &  & .994 &  & .403 &  & .455 &  &  .316 &  & .542  \\ \hline
\multicolumn{9}{c}{Supervise Fine-Tuned (SFT) models \textbf{with our method}.} \\ \hline
Ours+Intern2-VL 2B &  & .993 &  & .396 &  & .459 &  & .334 &  & .546  \\
Ours+Qwen2.5-VL 3B &  & .995 &  & .416 &  & .482 &  & .355 &  & .562  \\
Ours+Qwen2.5-VL 7B &  & \textbf{.995} &  & \textbf{.423} &  & \textbf{.490} &  & \textbf{.360} &  & \textbf{.567}  \\ \hline
\end{tabular}
\label{tab: mllm1}
\end{table}

\subsection{Ablation Study on Forgery Localization}
The ablation results are shown in Table~\ref{tab: ablmain}. 'w.o. MG' denotes the model without the Modal Gate and using frequency-vision fused features in encoder, it has an average IoU of .629 which is 8.5 points lower than Omni-IML of .714. This is because the frequency features in some samples are unstable, and without the Modal Gate to filter them out, these features introduce too much noise to the encoder and thus cause performance degradation. 'w.o. MG*' denotes the model without Modal Gate and using the pure vision modality, it has 5.5 points lower IoU than Omni-IML. This is because frequency domain modeling can also be helpful in some cases, especially when the tampered region is visually consistent (e.g. on document images). 'w.o. DWD' represents the model without the Dynamic Weight Decoder, it has 10.5 points lower IoU than Omni-IML. This is because the diversity of tampering features is too high for the encoder to learn them well, thus confusing the model, confirming the necessity of the proposed DWD for the generalist model. 'w.o. DW' is the model with the DWD structure but the filter weights in the decoder keep all the same for each input, it has 2.6 points lower IoU than Omni-IML, this verifies that the adaptive selection of optimal decoder weights for each sample can reduce confusion in joint training. 'w.o. AE' is the model without the proposed Anomaly Enhancement (AE) module, it has 2.9 points lower IoU than Omni-IML. This is because the proposed AE module can enhance the forged regions and suppress noise in the features. The model without any of the proposed modules serves as the 'Baseline' model, its IoU is 14.2 points lower than Omni-IML. These results confi the effectiveness of our methods.

\subsection{Experiments on Forgery Interpretation}
The experiments on interpretation task are conducted on our Omni-273k dataset. In Table~\ref{tab: mllm1}, we directly evaluate models with their whole output string for each sample using word-vectors cosine similarity~\cite{mikolov2017cossim}, ROUGE and BLEU, following the previous works~\cite{xu2024fakeshield, huang2024sida}. All the pre-trained models (the top group in Table~\ref{tab: mllm1}) show poor interpretation performance (e.g. all BLEU scores less than 0.1), confirming that the pre-trained LLMs are not naturally adequate to detect and explain image forgery. Fine-tuning the models with our dataset is necessary for them to achieve forgery interpretation capability. The models fine-tuned with our method (bottom group) consistently outperform those fine-tuned without our method (middle group) under all metrics, confirming the effectiveness of the reference visual prompt in our interpretation module.

Directly evaluating models on the unstructured string format as in previous works~\cite{xu2024fakeshield, huang2024sida} is coarse, as it can merely show the final result but cannot provide further insights behind the result. Moreover, all the common metrics are insufficient: cosine similarity is indiscriminate and ROUGE, BLEU can not recognize synonyms.
To provide better analysis, we convert the model output from JSON string to key-value pairs (Section 4.2) and perform fine-grained evaluation with them. As shown in Table~\ref{tab: mllm2}, we use OCR accuracy~\cite{zhang2019ic19} to evaluate the descriptions of tampered content and relative position for text images, and use Mean of ROUGE-L and BLEU (MRB) for non-text images. The description answers for absolute position (e.g. "Top left") and artifacts title (e.g. "Edge Artifacts") are close-ended, they can be regarded as single-choice and multi-choice tasks respectively, so we use accuracy to evaluate the ratio of matches. The detailed artifact description sentences are open-ended, and again we use the MRB score to evaluate them.

By comparing the model fine-tuned with and without our method, we can learn that our visual prompt improves model by reducing the misidentification of tampered regions, especially in multi-target, challenging scenarios such as documents. For example, for the Qwen2.5-VL 7B fine-tuned without our method, the description scores for tampered text, absolute and relative positions are $.312, .381, .429$ respectively on documents. These low scores mean that the model often incorrectly detect the tampered regions. In contrast, the same model fine-tuned with our method gets the three much higher scores of $.647, .569, .692$ respectively. This means our visual prompt assists the model to correctly identify the tampered region, and consequently the artifact description is much more accurate (e.g. $.681$ outperforms .521 in artifact description accuracy). Similar conclusion to other image types such as scene texts and faces. Our method is evaluated on various state-of-the-art LLMs and consistently yields obvious improvements (e.g. +21\% average score for Intern-VL 2B and + 16\% for Qwen2.5-VL 7B), demonstrating high generality and effectiveness. Model interpretation predictions are presented in the appendix.

\section{Conclusion}
In this paper, we propose Omni-IML, the first generalist model for Image Manipulation Localization (IML). Specifically, our generalist model achieves generalizable localization through multiple novel and effective modules, including a Modal Gate Encoder, a Dynamic Weight Decoder, and an Anomaly Enhancement module. 
Further, to enable language interpretation for image artifact, we construct a large-scale, comprehensive, high-quality dataset Omni-273k. The textual annotations in this dataset are automatically curated by a novel chain-of-thoughts pipeline, which significantly improves annotation quality. 
A simple-yet-effective interpretation module is also introduced to better interpret the tampered regions.
According to extensive experiments on four major IML tasks (natural IML, document IML, face IML, and scene text IML), our single model can achieve state-of-the-art performance simultaneously on all tasks. In-depth analysis are provided by our comprehensive ablation studies. 
We believe that our work can inspire future research and promote the real-world applications in unified image forensics.

\bibliographystyle{ACM-Reference-Format}
\bibliography{sample-base}


\begin{thebibliography}{49}

%%% the Web address.

\ifx \showCODEN    \undefined \def \showCODEN     #1{\unskip}     \fi
\ifx \showDOI      \undefined \def \showDOI       #1{#1}\fi
\ifx \showISBNx    \undefined \def \showISBNx     #1{\unskip}     \fi
\ifx \showISBNxiii \undefined \def \showISBNxiii  #1{\unskip}     \fi
\ifx \showISSN     \undefined \def \showISSN      #1{\unskip}     \fi
\ifx \showLCCN     \undefined \def \showLCCN      #1{\unskip}     \fi
\ifx \shownote     \undefined \def \shownote      #1{#1}          \fi
\ifx \showarticletitle \undefined \def \showarticletitle #1{#1}   \fi
\ifx \showURL      \undefined \def \showURL       {\relax}        \fi
% invisible to TeX
\providecommand\bibfield[2]{#2}
\providecommand\bibinfo[2]{#2}
\providecommand\natexlab[1]{#1}
\providecommand\showeprint[2][]{arXiv:#2}

\bibitem[{Alibaba Security}(2020)]%
        {sacp}
\bibfield{author}{\bibinfo{person}{{Alibaba Security}}.} \bibinfo{year}{2020}\natexlab{}.
\newblock \bibinfo{title}{Security AI Challenger Program}.
\newblock \bibinfo{howpublished}{\url{https://tianchi.aliyun.com/competition/entrance/531812/introduction}}.
\newblock


\bibitem[Bai et~al\mbox{.}(2025)]%
        {bai2025qwen2}
\bibfield{author}{\bibinfo{person}{Shuai Bai}, \bibinfo{person}{Keqin Chen}, \bibinfo{person}{Xuejing Liu}, \bibinfo{person}{Jialin Wang}, \bibinfo{person}{Wenbin Ge}, \bibinfo{person}{Sibo Song}, \bibinfo{person}{Kai Dang}, \bibinfo{person}{Peng Wang}, \bibinfo{person}{Shijie Wang}, \bibinfo{person}{Jun Tang}, {et~al\mbox{.}}} \bibinfo{year}{2025}\natexlab{}.
\newblock \showarticletitle{Qwen2. 5-vl technical report}.
\newblock \bibinfo{journal}{\emph{arXiv preprint arXiv:2502.13923}} (\bibinfo{year}{2025}).
\newblock


\bibitem[Bataineh et~al\mbox{.}(2011)]%
        {doc1}
\bibfield{author}{\bibinfo{person}{Bilal Bataineh}, \bibinfo{person}{Siti Norul Huda~Sheikh Abdullah}, {and} \bibinfo{person}{Khairudin Omar}.} \bibinfo{year}{2011}\natexlab{}.
\newblock \showarticletitle{A statistical global feature extraction method for optical font recognition}. In \bibinfo{booktitle}{\emph{Intelligent Information and Database Systems: Third International Conference, ACIIDS 2011, Daegu, Korea, April 20-22, 2011, Proceedings, Part I 3}}. Springer, \bibinfo{pages}{257--267}.
\newblock


\bibitem[Bi et~al\mbox{.}(2019)]%
        {bi2019rru}
\bibfield{author}{\bibinfo{person}{Xiuli Bi}, \bibinfo{person}{Yang Wei}, \bibinfo{person}{Bin Xiao}, {and} \bibinfo{person}{Weisheng Li}.} \bibinfo{year}{2019}\natexlab{}.
\newblock \showarticletitle{RRU-Net: The ringed residual U-Net for image splicing forgery detection}. In \bibinfo{booktitle}{\emph{Proceedings of the IEEE/CVF Conference on Computer Vision and Pattern Recognition Workshops}}. \bibinfo{pages}{0--0}.
\newblock


\bibitem[Chen et~al\mbox{.}(2024)]%
        {chen2024internvl}
\bibfield{author}{\bibinfo{person}{Zhe Chen}, \bibinfo{person}{Jiannan Wu}, \bibinfo{person}{Wenhai Wang}, \bibinfo{person}{Weijie Su}, \bibinfo{person}{Guo Chen}, \bibinfo{person}{Sen Xing}, \bibinfo{person}{Muyan Zhong}, \bibinfo{person}{Qinglong Zhang}, \bibinfo{person}{Xizhou Zhu}, \bibinfo{person}{Lewei Lu}, {et~al\mbox{.}}} \bibinfo{year}{2024}\natexlab{}.
\newblock \showarticletitle{Internvl: Scaling up vision foundation models and aligning for generic visual-linguistic tasks}. In \bibinfo{booktitle}{\emph{Proceedings of the IEEE/CVF conference on computer vision and pattern recognition}}. \bibinfo{pages}{24185--24198}.
\newblock


\bibitem[Dong et~al\mbox{.}(2022)]%
        {dong2022mvss}
\bibfield{author}{\bibinfo{person}{Chengbo Dong}, \bibinfo{person}{Xinru Chen}, \bibinfo{person}{Ruohan Hu}, \bibinfo{person}{Juan Cao}, {and} \bibinfo{person}{Xirong Li}.} \bibinfo{year}{2022}\natexlab{}.
\newblock \showarticletitle{Mvss-net: Multi-view multi-scale supervised networks for image manipulation detection}.
\newblock \bibinfo{journal}{\emph{IEEE Transactions on Pattern Analysis and Machine Intelligence}} \bibinfo{volume}{45}, \bibinfo{number}{3} (\bibinfo{year}{2022}), \bibinfo{pages}{3539--3553}.
\newblock


\bibitem[Dong et~al\mbox{.}(2013)]%
        {casia}
\bibfield{author}{\bibinfo{person}{Jing Dong}, \bibinfo{person}{Wei Wang}, {and} \bibinfo{person}{Tieniu Tan}.} \bibinfo{year}{2013}\natexlab{}.
\newblock \showarticletitle{CASIA Image Tampering Detection Evaluation Database}. In \bibinfo{booktitle}{\emph{2013 IEEE China Summit and International Conference on Signal and Information Processing}}. \bibinfo{pages}{422--426}.
\newblock
\urldef\tempurl%
\url{https://doi.org/10.1109/ChinaSIP.2013.6625374}
\showDOI{\tempurl}


\bibitem[Dong et~al\mbox{.}(2024)]%
        {dong2024robust}
\bibfield{author}{\bibinfo{person}{Renshuai~Liu Dong, Li}, \bibinfo{person}{Bowen Ma}, \bibinfo{person}{Wei Zhang}, \bibinfo{person}{Zhipeng Hu}, \bibinfo{person}{Changjie Fan}, \bibinfo{person}{Tangjie Lv}, \bibinfo{person}{Yu Ding}, {and} \bibinfo{person}{Xuan Cheng}.} \bibinfo{year}{2024}\natexlab{}.
\newblock \showarticletitle{Robust Text Image Tampering Localization via Forgery Traces Enhancement and Multiscale Attention}.
\newblock \bibinfo{journal}{\emph{IEEE Transactions on Consumer Electronics}} (\bibinfo{year}{2024}).
\newblock


\bibitem[Guan et~al\mbox{.}(2019)]%
        {nist16}
\bibfield{author}{\bibinfo{person}{Haiying Guan}, \bibinfo{person}{Mark Kozak}, \bibinfo{person}{Eric Robertson}, \bibinfo{person}{Yooyoung Lee}, \bibinfo{person}{Amy~N Yates}, \bibinfo{person}{Andrew Delgado}, \bibinfo{person}{Daniel Zhou}, \bibinfo{person}{Timothee Kheyrkhah}, \bibinfo{person}{Jeff Smith}, {and} \bibinfo{person}{Jonathan Fiscus}.} \bibinfo{year}{2019}\natexlab{}.
\newblock \showarticletitle{MFC datasets: Large-scale benchmark datasets for media forensic challenge evaluation}. In \bibinfo{booktitle}{\emph{2019 IEEE Winter Applications of Computer Vision Workshops (WACVW)}}. IEEE, \bibinfo{pages}{63--72}.
\newblock


\bibitem[Guillaro et~al\mbox{.}(2023)]%
        {guillaro2023trufor}
\bibfield{author}{\bibinfo{person}{Fabrizio Guillaro}, \bibinfo{person}{Davide Cozzolino}, \bibinfo{person}{Avneesh Sud}, \bibinfo{person}{Nicholas Dufour}, {and} \bibinfo{person}{Luisa Verdoliva}.} \bibinfo{year}{2023}\natexlab{}.
\newblock \showarticletitle{TruFor: Leveraging all-round clues for trustworthy image forgery detection and localization}. In \bibinfo{booktitle}{\emph{Proceedings of the IEEE/CVF Conference on Computer Vision and Pattern Recognition}}. \bibinfo{pages}{20606--20615}.
\newblock


\bibitem[Guo et~al\mbox{.}(2023)]%
        {hifi_net_xiaoguo}
\bibfield{author}{\bibinfo{person}{Xiao Guo}, \bibinfo{person}{Xiaohong Liu}, \bibinfo{person}{Zhiyuan Ren}, \bibinfo{person}{Steven Grosz}, \bibinfo{person}{Iacopo Masi}, {and} \bibinfo{person}{Xiaoming Liu}.} \bibinfo{year}{2023}\natexlab{}.
\newblock \showarticletitle{Hierarchical Fine-Grained Image Forgery Detection and Localization}. In \bibinfo{booktitle}{\emph{CVPR}}.
\newblock


\bibitem[Heuer et~al\mbox{.}(2021)]%
        {heuer2021multitask}
\bibfield{author}{\bibinfo{person}{Falk Heuer}, \bibinfo{person}{Sven Mantowsky}, \bibinfo{person}{Saqib Bukhari}, {and} \bibinfo{person}{Georg Schneider}.} \bibinfo{year}{2021}\natexlab{}.
\newblock \showarticletitle{Multitask-centernet (mcn): Efficient and diverse multitask learning using an anchor free approach}. In \bibinfo{booktitle}{\emph{Proceedings of the IEEE/CVF International conference on computer vision}}. \bibinfo{pages}{997--1005}.
\newblock


\bibitem[Hu et~al\mbox{.}(2022)]%
        {hu2022lora}
\bibfield{author}{\bibinfo{person}{Edward~J Hu}, \bibinfo{person}{Yelong Shen}, \bibinfo{person}{Phillip Wallis}, \bibinfo{person}{Zeyuan Allen-Zhu}, \bibinfo{person}{Yuanzhi Li}, \bibinfo{person}{Shean Wang}, \bibinfo{person}{Lu Wang}, \bibinfo{person}{Weizhu Chen}, {et~al\mbox{.}}} \bibinfo{year}{2022}\natexlab{}.
\newblock \showarticletitle{Lora: Low-rank adaptation of large language models.}
\newblock \bibinfo{journal}{\emph{ICLR}} \bibinfo{volume}{1}, \bibinfo{number}{2} (\bibinfo{year}{2022}), \bibinfo{pages}{3}.
\newblock


\bibitem[Huang et~al\mbox{.}(2024a)]%
        {huang2024sida}
\bibfield{author}{\bibinfo{person}{Zhenglin Huang}, \bibinfo{person}{Jinwei Hu}, \bibinfo{person}{Xiangtai Li}, \bibinfo{person}{Yiwei He}, \bibinfo{person}{Xingyu Zhao}, \bibinfo{person}{Bei Peng}, \bibinfo{person}{Baoyuan Wu}, \bibinfo{person}{Xiaowei Huang}, {and} \bibinfo{person}{Guangliang Cheng}.} \bibinfo{year}{2024}\natexlab{a}.
\newblock \showarticletitle{SIDA: Social Media Image Deepfake Detection, Localization and Explanation with Large Multimodal Model}.
\newblock \bibinfo{journal}{\emph{arXiv preprint arXiv:2412.04292}} (\bibinfo{year}{2024}).
\newblock


\bibitem[Huang et~al\mbox{.}(2024b)]%
        {huang2024ffaa}
\bibfield{author}{\bibinfo{person}{Zhengchao Huang}, \bibinfo{person}{Bin Xia}, \bibinfo{person}{Zicheng Lin}, \bibinfo{person}{Zhun Mou}, \bibinfo{person}{Wenming Yang}, {and} \bibinfo{person}{Jiaya Jia}.} \bibinfo{year}{2024}\natexlab{b}.
\newblock \showarticletitle{Ffaa: Multimodal large language model based explainable open-world face forgery analysis assistant}.
\newblock \bibinfo{journal}{\emph{arXiv preprint arXiv:2408.10072}} (\bibinfo{year}{2024}).
\newblock


\bibitem[Jia et~al\mbox{.}(2023)]%
        {autosplice}
\bibfield{author}{\bibinfo{person}{Shan Jia}, \bibinfo{person}{Mingzhen Huang}, \bibinfo{person}{Zhou Zhou}, \bibinfo{person}{Yan Ju}, \bibinfo{person}{Jialing Cai}, {and} \bibinfo{person}{Siwei Lyu}.} \bibinfo{year}{2023}\natexlab{}.
\newblock \showarticletitle{AutoSplice: A Text-Prompt Manipulated Image Dataset for Media Forensics}. In \bibinfo{booktitle}{\emph{Proceedings of the IEEE/CVF Conference on Computer Vision and Pattern Recognition (CVPR) Workshops}}. \bibinfo{pages}{893--903}.
\newblock


\bibitem[Kwon et~al\mbox{.}(2022)]%
        {kwon2022learning}
\bibfield{author}{\bibinfo{person}{Myung-Joon Kwon}, \bibinfo{person}{Seung-Hun Nam}, \bibinfo{person}{In-Jae Yu}, \bibinfo{person}{Heung-Kyu Lee}, {and} \bibinfo{person}{Changick Kim}.} \bibinfo{year}{2022}\natexlab{}.
\newblock \showarticletitle{Learning JPEG compression artifacts for image manipulation detection and localization}.
\newblock \bibinfo{journal}{\emph{International Journal of Computer Vision}} \bibinfo{volume}{130}, \bibinfo{number}{8} (\bibinfo{year}{2022}), \bibinfo{pages}{1875--1895}.
\newblock


\bibitem[Lai et~al\mbox{.}(2024)]%
        {lai2024lisa}
\bibfield{author}{\bibinfo{person}{Xin Lai}, \bibinfo{person}{Zhuotao Tian}, \bibinfo{person}{Yukang Chen}, \bibinfo{person}{Yanwei Li}, \bibinfo{person}{Yuhui Yuan}, \bibinfo{person}{Shu Liu}, {and} \bibinfo{person}{Jiaya Jia}.} \bibinfo{year}{2024}\natexlab{}.
\newblock \showarticletitle{Lisa: Reasoning segmentation via large language model}. In \bibinfo{booktitle}{\emph{Proceedings of the IEEE/CVF Conference on Computer Vision and Pattern Recognition}}. \bibinfo{pages}{9579--9589}.
\newblock


\bibitem[Le et~al\mbox{.}(2021)]%
        {le2021openforensics}
\bibfield{author}{\bibinfo{person}{Trung-Nghia Le}, \bibinfo{person}{Huy~H Nguyen}, \bibinfo{person}{Junichi Yamagishi}, {and} \bibinfo{person}{Isao Echizen}.} \bibinfo{year}{2021}\natexlab{}.
\newblock \showarticletitle{Openforensics: Large-scale challenging dataset for multi-face forgery detection and segmentation in-the-wild}. In \bibinfo{booktitle}{\emph{Proceedings of the IEEE/CVF international conference on computer vision}}. \bibinfo{pages}{10117--10127}.
\newblock


\bibitem[Li et~al\mbox{.}(2024)]%
        {Li_2024_CVPR}
\bibfield{author}{\bibinfo{person}{Shuaibo Li}, \bibinfo{person}{Wei Ma}, \bibinfo{person}{Jianwei Guo}, \bibinfo{person}{Shibiao Xu}, \bibinfo{person}{Benchong Li}, {and} \bibinfo{person}{Xiaopeng Zhang}.} \bibinfo{year}{2024}\natexlab{}.
\newblock \showarticletitle{UnionFormer: Unified-Learning Transformer with Multi-View Representation for Image Manipulation Detection and Localization}. In \bibinfo{booktitle}{\emph{Proceedings of the IEEE/CVF Conference on Computer Vision and Pattern Recognition (CVPR)}}. \bibinfo{pages}{12523--12533}.
\newblock


\bibitem[Lin et~al\mbox{.}(2017)]%
        {lin2017fpn}
\bibfield{author}{\bibinfo{person}{Tsung-Yi Lin}, \bibinfo{person}{Piotr Doll{\'a}r}, \bibinfo{person}{Ross Girshick}, \bibinfo{person}{Kaiming He}, \bibinfo{person}{Bharath Hariharan}, {and} \bibinfo{person}{Serge Belongie}.} \bibinfo{year}{2017}\natexlab{}.
\newblock \showarticletitle{Feature pyramid networks for object detection}. In \bibinfo{booktitle}{\emph{Proceedings of the IEEE conference on computer vision and pattern recognition}}. \bibinfo{pages}{2117--2125}.
\newblock


\bibitem[Lin et~al\mbox{.}(2014)]%
        {lin2014microsoft}
\bibfield{author}{\bibinfo{person}{Tsung-Yi Lin}, \bibinfo{person}{Michael Maire}, \bibinfo{person}{Serge Belongie}, \bibinfo{person}{James Hays}, \bibinfo{person}{Pietro Perona}, \bibinfo{person}{Deva Ramanan}, \bibinfo{person}{Piotr Doll{\'a}r}, {and} \bibinfo{person}{C~Lawrence Zitnick}.} \bibinfo{year}{2014}\natexlab{}.
\newblock \showarticletitle{Microsoft coco: Common objects in context}. In \bibinfo{booktitle}{\emph{Computer Vision--ECCV 2014: 13th European Conference, Zurich, Switzerland, September 6-12, 2014, Proceedings, Part V 13}}. Springer, \bibinfo{pages}{740--755}.
\newblock


\bibitem[Liu et~al\mbox{.}(2023)]%
        {liu2023explicit}
\bibfield{author}{\bibinfo{person}{Weihuang Liu}, \bibinfo{person}{Xi Shen}, \bibinfo{person}{Chi-Man Pun}, {and} \bibinfo{person}{Xiaodong Cun}.} \bibinfo{year}{2023}\natexlab{}.
\newblock \showarticletitle{Explicit visual prompting for low-level structure segmentations}. In \bibinfo{booktitle}{\emph{Proceedings of the IEEE/CVF Conference on Computer Vision and Pattern Recognition}}. \bibinfo{pages}{19434--19445}.
\newblock


\bibitem[Liu et~al\mbox{.}(2022a)]%
        {liu2022pscc}
\bibfield{author}{\bibinfo{person}{Xiaohong Liu}, \bibinfo{person}{Yaojie Liu}, \bibinfo{person}{Jun Chen}, {and} \bibinfo{person}{Xiaoming Liu}.} \bibinfo{year}{2022}\natexlab{a}.
\newblock \showarticletitle{PSCC-Net: Progressive spatio-channel correlation network for image manipulation detection and localization}.
\newblock \bibinfo{journal}{\emph{IEEE Transactions on Circuits and Systems for Video Technology}} \bibinfo{volume}{32}, \bibinfo{number}{11} (\bibinfo{year}{2022}), \bibinfo{pages}{7505--7517}.
\newblock


\bibitem[Liu et~al\mbox{.}(2024)]%
        {liu2024hfnet}
\bibfield{author}{\bibinfo{person}{Yang Liu}, \bibinfo{person}{Xiaofei Li}, \bibinfo{person}{Jun Zhang}, \bibinfo{person}{Shengze Hu}, {and} \bibinfo{person}{Jun Lei}.} \bibinfo{year}{2024}\natexlab{}.
\newblock \showarticletitle{DA-HFNet: Progressive Fine-Grained Forgery Image Detection and Localization Based on Dual Attention}.
\newblock \bibinfo{journal}{\emph{arXiv preprint arXiv:2406.01489}} (\bibinfo{year}{2024}).
\newblock


\bibitem[Liu et~al\mbox{.}(2022b)]%
        {liu2022convnet}
\bibfield{author}{\bibinfo{person}{Zhuang Liu}, \bibinfo{person}{Hanzi Mao}, \bibinfo{person}{Chao-Yuan Wu}, \bibinfo{person}{Christoph Feichtenhofer}, \bibinfo{person}{Trevor Darrell}, {and} \bibinfo{person}{Saining Xie}.} \bibinfo{year}{2022}\natexlab{b}.
\newblock \showarticletitle{A convnet for the 2020s}. In \bibinfo{booktitle}{\emph{Proceedings of the IEEE/CVF conference on computer vision and pattern recognition}}. \bibinfo{pages}{11976--11986}.
\newblock


\bibitem[Loshchilov and Hutter(2017)]%
        {adamw}
\bibfield{author}{\bibinfo{person}{Ilya Loshchilov} {and} \bibinfo{person}{Frank Hutter}.} \bibinfo{year}{2017}\natexlab{}.
\newblock \showarticletitle{Decoupled weight decay regularization}.
\newblock \bibinfo{journal}{\emph{arXiv preprint arXiv:1711.05101}} (\bibinfo{year}{2017}).
\newblock


\bibitem[Lu et~al\mbox{.}(2024)]%
        {lu2024deepseek}
\bibfield{author}{\bibinfo{person}{Haoyu Lu}, \bibinfo{person}{Wen Liu}, \bibinfo{person}{Bo Zhang}, \bibinfo{person}{Bingxuan Wang}, \bibinfo{person}{Kai Dong}, \bibinfo{person}{Bo Liu}, \bibinfo{person}{Jingxiang Sun}, \bibinfo{person}{Tongzheng Ren}, \bibinfo{person}{Zhuoshu Li}, \bibinfo{person}{Hao Yang}, {et~al\mbox{.}}} \bibinfo{year}{2024}\natexlab{}.
\newblock \showarticletitle{Deepseek-vl: towards real-world vision-language understanding}.
\newblock \bibinfo{journal}{\emph{arXiv preprint arXiv:2403.05525}} (\bibinfo{year}{2024}).
\newblock


\bibitem[Ma et~al\mbox{.}(2024)]%
        {ma2024benco}
\bibfield{author}{\bibinfo{person}{Xiaochen Ma}, \bibinfo{person}{Xuekang Zhu}, \bibinfo{person}{Lei Su}, \bibinfo{person}{Bo Du}, \bibinfo{person}{Zhuohang Jiang}, \bibinfo{person}{Bingkui Tong}, \bibinfo{person}{Zeyu Lei}, \bibinfo{person}{Xinyu Yang}, \bibinfo{person}{Chi-Man Pun}, \bibinfo{person}{Jiancheng Lv}, {et~al\mbox{.}}} \bibinfo{year}{2024}\natexlab{}.
\newblock \showarticletitle{Imdl-benco: A comprehensive benchmark and codebase for image manipulation detection \& localization}.
\newblock \bibinfo{journal}{\emph{Advances in Neural Information Processing Systems}}  \bibinfo{volume}{37} (\bibinfo{year}{2024}), \bibinfo{pages}{134591--134613}.
\newblock


\bibitem[Mikolov et~al\mbox{.}(2017)]%
        {mikolov2017cossim}
\bibfield{author}{\bibinfo{person}{Tomas Mikolov}, \bibinfo{person}{Edouard Grave}, \bibinfo{person}{Piotr Bojanowski}, \bibinfo{person}{Christian Puhrsch}, {and} \bibinfo{person}{Armand Joulin}.} \bibinfo{year}{2017}\natexlab{}.
\newblock \showarticletitle{Advances in pre-training distributed word representations}.
\newblock \bibinfo{journal}{\emph{arXiv preprint arXiv:1712.09405}} (\bibinfo{year}{2017}).
\newblock


\bibitem[Novozamsky et~al\mbox{.}(2020)]%
        {imd20}
\bibfield{author}{\bibinfo{person}{Adam Novozamsky}, \bibinfo{person}{Babak Mahdian}, {and} \bibinfo{person}{Stanislav Saic}.} \bibinfo{year}{2020}\natexlab{}.
\newblock \showarticletitle{IMD2020: A Large-Scale Annotated Dataset Tailored for Detecting Manipulated Images}. In \bibinfo{booktitle}{\emph{Proceedings of the IEEE/CVF Winter Conference on Applications of Computer Vision (WACV) Workshops}}.
\newblock


\bibitem[Qu et~al\mbox{.}(2023)]%
        {CVPR2023DocTamper}
\bibfield{author}{\bibinfo{person}{Chenfan Qu}, \bibinfo{person}{Chongyu Liu}, \bibinfo{person}{Yuliang Liu}, \bibinfo{person}{Xinhong Chen}, \bibinfo{person}{Dezhi Peng}, \bibinfo{person}{Fengjun Guo}, {and} \bibinfo{person}{Lianwen Jin}.} \bibinfo{year}{2023}\natexlab{}.
\newblock \showarticletitle{Towards robust tampered text detection in document image: new dataset and new solution}. In \bibinfo{booktitle}{\emph{Proceedings of the IEEE/CVF Conference on Computer Vision and Pattern Recognition}}. \bibinfo{pages}{5937--5946}.
\newblock


\bibitem[Qu et~al\mbox{.}(2025)]%
        {ostf}
\bibfield{author}{\bibinfo{person}{Chenfan Qu}, \bibinfo{person}{Yiwu Zhong}, \bibinfo{person}{Fengjun Guo}, {and} \bibinfo{person}{Lianwen Jin}.} \bibinfo{year}{2025}\natexlab{}.
\newblock \showarticletitle{Revisiting Tampered Scene Text Detection in the Era of Generative AI}. In \bibinfo{booktitle}{\emph{AAAI Conference on Artificial Intelligence}}. AAAI, \bibinfo{pages}{568--584}.
\newblock


\bibitem[Qu et~al\mbox{.}(2024)]%
        {Qu_2024_CVPR}
\bibfield{author}{\bibinfo{person}{Chenfan Qu}, \bibinfo{person}{Yiwu Zhong}, \bibinfo{person}{Chongyu Liu}, \bibinfo{person}{Guitao Xu}, \bibinfo{person}{Dezhi Peng}, \bibinfo{person}{Fengjun Guo}, {and} \bibinfo{person}{Lianwen Jin}.} \bibinfo{year}{2024}\natexlab{}.
\newblock \showarticletitle{Towards Modern Image Manipulation Localization: A Large-Scale Dataset and Novel Methods}. In \bibinfo{booktitle}{\emph{Proceedings of the IEEE/CVF Conference on Computer Vision and Pattern Recognition (CVPR)}}. \bibinfo{pages}{10781--10790}.
\newblock


\bibitem[Ren et~al\mbox{.}(2015)]%
        {ren2015fasterrcnn}
\bibfield{author}{\bibinfo{person}{Shaoqing Ren}, \bibinfo{person}{Kaiming He}, \bibinfo{person}{Ross Girshick}, {and} \bibinfo{person}{Jian Sun}.} \bibinfo{year}{2015}\natexlab{}.
\newblock \showarticletitle{Faster r-cnn: Towards real-time object detection with region proposal networks}.
\newblock \bibinfo{journal}{\emph{Advances in neural information processing systems}}  \bibinfo{volume}{28} (\bibinfo{year}{2015}).
\newblock


\bibitem[Sun et~al\mbox{.}(2023)]%
        {sun2023safl}
\bibfield{author}{\bibinfo{person}{Zhihao Sun}, \bibinfo{person}{Haoran Jiang}, \bibinfo{person}{Danding Wang}, \bibinfo{person}{Xirong Li}, {and} \bibinfo{person}{Juan Cao}.} \bibinfo{year}{2023}\natexlab{}.
\newblock \showarticletitle{SAFL-Net: Semantic-Agnostic Feature Learning Network with Auxiliary Plugins for Image Manipulation Detection}. In \bibinfo{booktitle}{\emph{Proceedings of the IEEE/CVF International Conference on Computer Vision}}. \bibinfo{pages}{22424--22433}.
\newblock


\bibitem[Van~Beusekom et~al\mbox{.}(2013)]%
        {doc2}
\bibfield{author}{\bibinfo{person}{Joost Van~Beusekom}, \bibinfo{person}{Faisal Shafait}, {and} \bibinfo{person}{Thomas~M Breuel}.} \bibinfo{year}{2013}\natexlab{}.
\newblock \showarticletitle{Text-line examination for document forgery detection}.
\newblock \bibinfo{journal}{\emph{International Journal on Document Analysis and Recognition (IJDAR)}}  \bibinfo{volume}{16} (\bibinfo{year}{2013}), \bibinfo{pages}{189--207}.
\newblock


\bibitem[Wang et~al\mbox{.}(2022a)]%
        {wang2022objectformer}
\bibfield{author}{\bibinfo{person}{Junke Wang}, \bibinfo{person}{Zuxuan Wu}, \bibinfo{person}{Jingjing Chen}, \bibinfo{person}{Xintong Han}, \bibinfo{person}{Abhinav Shrivastava}, \bibinfo{person}{Ser-Nam Lim}, {and} \bibinfo{person}{Yu-Gang Jiang}.} \bibinfo{year}{2022}\natexlab{a}.
\newblock \showarticletitle{Objectformer for image manipulation detection and localization}. In \bibinfo{booktitle}{\emph{Proceedings of the IEEE/CVF Conference on Computer Vision and Pattern Recognition}}. \bibinfo{pages}{2364--2373}.
\newblock


\bibitem[Wang et~al\mbox{.}(2022b)]%
        {wang2022tic}
\bibfield{author}{\bibinfo{person}{Yuxin Wang}, \bibinfo{person}{Hongtao Xie}, \bibinfo{person}{Mengting Xing}, \bibinfo{person}{Jing Wang}, \bibinfo{person}{Shenggao Zhu}, {and} \bibinfo{person}{Yongdong Zhang}.} \bibinfo{year}{2022}\natexlab{b}.
\newblock \showarticletitle{Detecting tampered scene text in the wild}. In \bibinfo{booktitle}{\emph{European Conference on Computer Vision}}. Springer, \bibinfo{pages}{215--232}.
\newblock


\bibitem[Wen et~al\mbox{.}(2016)]%
        {coverage}
\bibfield{author}{\bibinfo{person}{Bihan Wen}, \bibinfo{person}{Ye Zhu}, \bibinfo{person}{Ramanathan Subramanian}, \bibinfo{person}{Tian-Tsong Ng}, \bibinfo{person}{Xuanjing Shen}, {and} \bibinfo{person}{Stefan Winkler}.} \bibinfo{year}{2016}\natexlab{}.
\newblock \showarticletitle{COVERAGE — A novel database for copy-move forgery detection}. In \bibinfo{booktitle}{\emph{2016 IEEE International Conference on Image Processing (ICIP)}}. \bibinfo{pages}{161--165}.
\newblock
\urldef\tempurl%
\url{https://doi.org/10.1109/ICIP.2016.7532339}
\showDOI{\tempurl}


\bibitem[Wu et~al\mbox{.}(2022)]%
        {wu2022robust}
\bibfield{author}{\bibinfo{person}{Haiwei Wu}, \bibinfo{person}{Jiantao Zhou}, \bibinfo{person}{Jinyu Tian}, {and} \bibinfo{person}{Jun Liu}.} \bibinfo{year}{2022}\natexlab{}.
\newblock \showarticletitle{Robust image forgery detection over online social network shared images}. In \bibinfo{booktitle}{\emph{Proceedings of the IEEE/CVF Conference on Computer Vision and Pattern Recognition}}. \bibinfo{pages}{13440--13449}.
\newblock


\bibitem[Wu et~al\mbox{.}(2019)]%
        {wu2019mantra}
\bibfield{author}{\bibinfo{person}{Yue Wu}, \bibinfo{person}{Zuxuan Wu}, \bibinfo{person}{Jingjing Chen}, \bibinfo{person}{Xintong Han}, \bibinfo{person}{Abhinav Shrivastava}, \bibinfo{person}{Ser-Nam Lim}, {and} \bibinfo{person}{Yu-Gang Jiang}.} \bibinfo{year}{2019}\natexlab{}.
\newblock \showarticletitle{Mantra-net: Manipulation tracing network for detection and localization of image forgeries with anomalous features}. In \bibinfo{booktitle}{\emph{Proceedings of the IEEE/CVF Conference on Computer Vision and Pattern Recognition}}. \bibinfo{pages}{9543--9552}.
\newblock


\bibitem[Xia et~al\mbox{.}(2024)]%
        {xia2024gsva}
\bibfield{author}{\bibinfo{person}{Zhuofan Xia}, \bibinfo{person}{Dongchen Han}, \bibinfo{person}{Yizeng Han}, \bibinfo{person}{Xuran Pan}, \bibinfo{person}{Shiji Song}, {and} \bibinfo{person}{Gao Huang}.} \bibinfo{year}{2024}\natexlab{}.
\newblock \showarticletitle{Gsva: Generalized segmentation via multimodal large language models}. In \bibinfo{booktitle}{\emph{Proceedings of the IEEE/CVF Conference on Computer Vision and Pattern Recognition}}. \bibinfo{pages}{3858--3869}.
\newblock


\bibitem[Xu et~al\mbox{.}(2024)]%
        {xu2024fakeshield}
\bibfield{author}{\bibinfo{person}{Zhipei Xu}, \bibinfo{person}{Xuanyu Zhang}, \bibinfo{person}{Runyi Li}, \bibinfo{person}{Zecheng Tang}, \bibinfo{person}{Qing Huang}, {and} \bibinfo{person}{Jian Zhang}.} \bibinfo{year}{2024}\natexlab{}.
\newblock \showarticletitle{Fakeshield: Explainable image forgery detection and localization via multi-modal large language models}.
\newblock \bibinfo{journal}{\emph{arXiv preprint arXiv:2410.02761}} (\bibinfo{year}{2024}).
\newblock


\bibitem[Yao et~al\mbox{.}(2024)]%
        {yao2024minicpm}
\bibfield{author}{\bibinfo{person}{Yuan Yao}, \bibinfo{person}{Tianyu Yu}, \bibinfo{person}{Ao Zhang}, \bibinfo{person}{Chongyi Wang}, \bibinfo{person}{Junbo Cui}, \bibinfo{person}{Hongji Zhu}, \bibinfo{person}{Tianchi Cai}, \bibinfo{person}{Haoyu Li}, \bibinfo{person}{Weilin Zhao}, \bibinfo{person}{Zhihui He}, {et~al\mbox{.}}} \bibinfo{year}{2024}\natexlab{}.
\newblock \showarticletitle{Minicpm-v: A gpt-4v level mllm on your phone}.
\newblock \bibinfo{journal}{\emph{arXiv preprint arXiv:2408.01800}} (\bibinfo{year}{2024}).
\newblock


\bibitem[Yu et~al\mbox{.}(2024)]%
        {Yu_2024_CVPR}
\bibfield{author}{\bibinfo{person}{Zeqin Yu}, \bibinfo{person}{Jiangqun Ni}, \bibinfo{person}{Yuzhen Lin}, \bibinfo{person}{Haoyi Deng}, {and} \bibinfo{person}{Bin Li}.} \bibinfo{year}{2024}\natexlab{}.
\newblock \showarticletitle{DiffForensics: Leveraging Diffusion Prior to Image Forgery Detection and Localization}. In \bibinfo{booktitle}{\emph{Proceedings of the IEEE/CVF Conference on Computer Vision and Pattern Recognition (CVPR)}}. \bibinfo{pages}{12765--12774}.
\newblock


\bibitem[Zhang et~al\mbox{.}(2019)]%
        {zhang2019ic19}
\bibfield{author}{\bibinfo{person}{Rui Zhang}, \bibinfo{person}{Yongsheng Zhou}, \bibinfo{person}{Qianyi Jiang}, \bibinfo{person}{Qi Song}, \bibinfo{person}{Nan Li}, \bibinfo{person}{Kai Zhou}, \bibinfo{person}{Lei Wang}, \bibinfo{person}{Dong Wang}, \bibinfo{person}{Minghui Liao}, \bibinfo{person}{Mingkun Yang}, {et~al\mbox{.}}} \bibinfo{year}{2019}\natexlab{}.
\newblock \showarticletitle{Icdar 2019 robust reading challenge on reading chinese text on signboard}. In \bibinfo{booktitle}{\emph{2019 international conference on document analysis and recognition (ICDAR)}}. IEEE, \bibinfo{pages}{1577--1581}.
\newblock


\bibitem[Zhao et~al\mbox{.}(2017)]%
        {zhao2017pyramid}
\bibfield{author}{\bibinfo{person}{Hengshuang Zhao}, \bibinfo{person}{Jianping Shi}, \bibinfo{person}{Xiaojuan Qi}, \bibinfo{person}{Xiaogang Wang}, {and} \bibinfo{person}{Jiaya Jia}.} \bibinfo{year}{2017}\natexlab{}.
\newblock \showarticletitle{Pyramid scene parsing network}. In \bibinfo{booktitle}{\emph{Proceedings of the IEEE conference on computer vision and pattern recognition}}. \bibinfo{pages}{2881--2890}.
\newblock


\bibitem[Zhou et~al\mbox{.}(2017)]%
        {ade20k}
\bibfield{author}{\bibinfo{person}{Bolei Zhou}, \bibinfo{person}{Hang Zhao}, \bibinfo{person}{Xavier Puig}, \bibinfo{person}{Sanja Fidler}, \bibinfo{person}{Adela Barriuso}, {and} \bibinfo{person}{Antonio Torralba}.} \bibinfo{year}{2017}\natexlab{}.
\newblock \showarticletitle{Scene parsing through ade20k dataset}. In \bibinfo{booktitle}{\emph{Proceedings of the IEEE conference on computer vision and pattern recognition}}. \bibinfo{pages}{633--641}.
\newblock


\end{thebibliography}

\end{document}